\documentclass[journal]{IEEEtran}
\usepackage{cite}
\ifCLASSINFOpdf
   \usepackage[pdftex]{graphicx}
\else
\fi
\usepackage{amsmath}
\usepackage{array}
\ifCLASSOPTIONcompsoc
 \usepackage[caption=false,font=normalsize,labelfont=sf,textfont=sf]{subfig}
\else
 \usepackage[caption=false,font=footnotesize]{subfig}
\fi
\usepackage{url}
\allowdisplaybreaks
\usepackage{amsfonts}
\usepackage{amssymb}
\usepackage{hyperref}
\usepackage[ruled,norelsize,vlined,linesnumbered]{algorithm2e}
\usepackage{float}
\usepackage[misc,geometry]{ifsym}
\usepackage{multirow}
\usepackage{makecell}
\usepackage{doi}
\usepackage{booktabs}
\usepackage{color}
\usepackage{ulem}
\usepackage[font=small,labelfont=bf]{caption}
\usepackage{algpseudocode}

\newtheorem{assumption}{Assumption}
\newtheorem{proposition}{Proposition}
\newtheorem{definition}{Definition}

\newcommand{\NAME}{\textsc{HFedMS}}

\newcommand{\removelatexerror}{\let\@latex@error\@gobble}


\begin{document}
\title{HFedMS: Heterogeneous Federated Learning with Memorable Data Semantics in Industrial Metaverse}

% \author{Anonymous Authors}
\author{
Shenglai~Zeng$^1$,
Zonghang~Li$^1$,
Hongfang~Yu (\Letter),
Zhihao~Zhang,
Long~Luo, \\
Bo~Li,~\IEEEmembership{Fellow~IEEE},
Dusit~Niyato,~\IEEEmembership{Fellow~IEEE}
\thanks{Shenglai Zeng and Zhihao Zhang are with Yingcai Honors College and School of Computer Science and Engineering, University of Electronic Science and Technology of China, Chengdu, China.}
\thanks{Zonghang Li, Hongfang Yu, and Long Luo are with School of Information and Communication Engineering, University of Electronic Science and Technology of China, Chengdu, China.}
\thanks{Bo Li is with Department of Computer Science and Engineering, Hong Kong University of Science and Technology, Hong Kong, China.}
\thanks{Dusit Niyato is with School of Computer Science and Engineering, Nanyang Technological University, Singapore.}
\thanks{Shenglai Zeng and Zonghang Li are of equal contributions. The corresponding author: Hongfang Yu. (Email: \url{yuhfnetworklab@gmail.com)}}
\thanks{This work was supported in part by National Natural Science Foundation of China (62102066), PCL Future Greater-Bay Area Network Facilities for Large-Scale Experiments and Applications (LZC0019), Open Research Projects of Zhejiang Lab (2022QA0AB02). This work was done in part at Nanyang Technological University, Singapore.}
}

\maketitle


% Abstract & Keywords
\begin{abstract}
Federated Learning (FL), as a rapidly evolving privacy-preserving collaborative machine learning paradigm, is a promising approach to enable edge intelligence in the emerging Industrial Metaverse. Even though many successful use cases have proved the feasibility of FL in theory, in the industrial practice of Metaverse, the problems of non-independent and identically distributed (non-i.i.d.) data, learning forgetting caused by streaming industrial data, and scarce communication bandwidth remain key barriers to realize practical FL. Facing the above three challenges simultaneously, this paper presents a high-performance and efficient system named \NAME~for incorporating practical FL into Industrial Metaverse. \NAME~reduces data heterogeneity through dynamic grouping and training mode conversion (\textit{Dynamic Sequential-to-Parallel Training, STP}). Then, it compensates for the forgotten knowledge by fusing compressed historical data semantics and calibrates classifier parameters (\textit{Semantic Compression and Compensation, SCC}). Finally, the network parameters of the feature extractor and classifier are synchronized in different frequencies (\textit{Layer-wise Alternative Synchronization Protocol, LASP}) to reduce communication costs. These techniques make FL more adaptable to the heterogeneous streaming data continuously generated by industrial equipment, and are also more efficient in communication than traditional methods (e.g., Federated Averaging). Extensive experiments have been conducted on the streamed non-i.i.d. FEMNIST dataset using 368 simulated devices. Numerical results show that \NAME~improves the classification accuracy by at least 6.4\% compared with 8 benchmarks and saves both the overall runtime and transfer bytes by up to 98\%, proving its superiority in precision and efficiency.

\end{abstract}

% Main Text
% \fi
\IEEEpeerreviewmaketitle

\section{Introduction}
Known as ``the successor of mobile Internet", the concept of Metaverse\cite{xu2022full} has attracted growing attention in both academia and industry. As a future interaction paradigm that requires a rich variety of enabling technologies, Metaverse would revolutionize many domains and their applications work. One possible application could be in the field of the Smart Industry, also known as the Industrial Metaverse. In fact, the Industrial Metaverse is said to be the field closest to realizing Metaverse, and there are already some practices in real factories. For example, Nvidia Omniverse 6 allows BMW to integrate its brick-and-mortar car factory with Virtual Reality (VR), Artificial Intelligence (AI), and robotics to improve its operation precision and flexibility. 

\begin{figure}[t]
\centering
\includegraphics[width=0.5\textwidth]{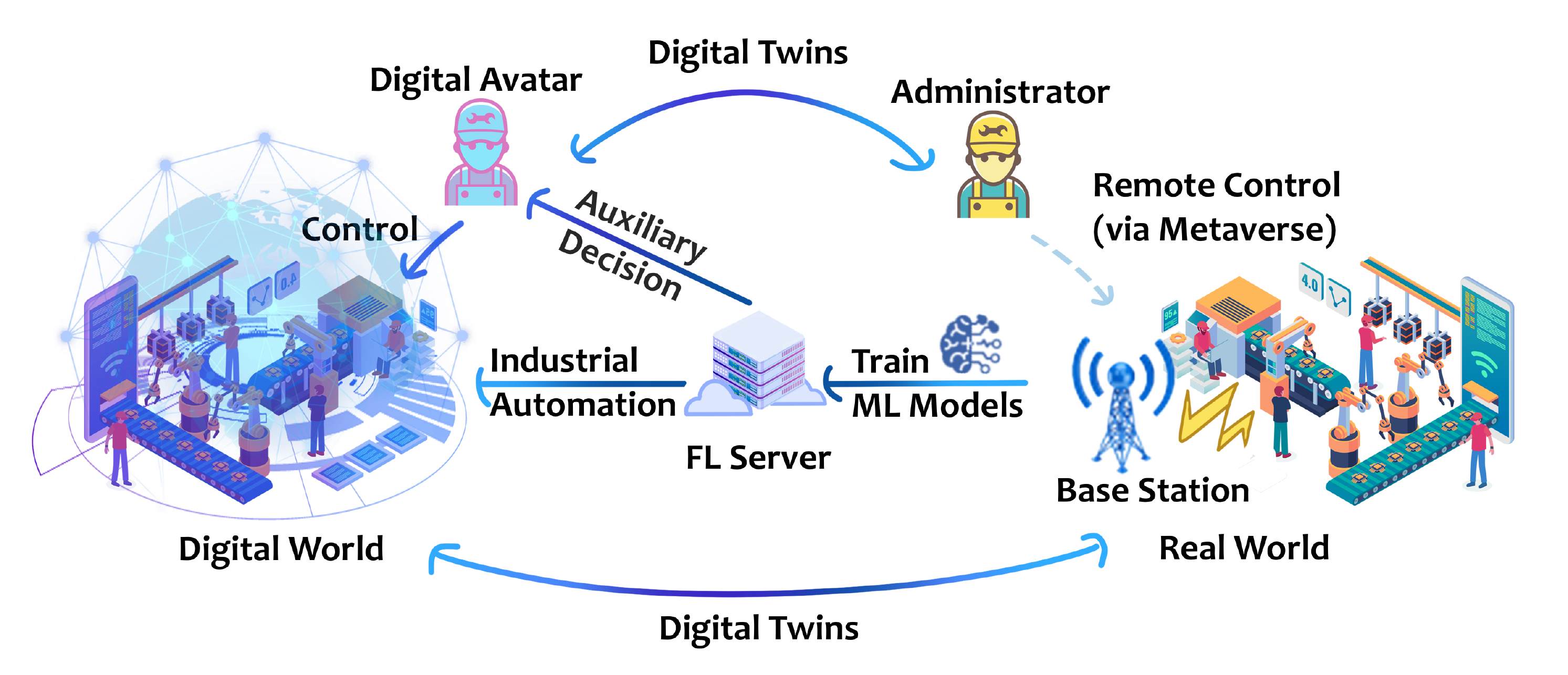}
\caption{Illustration of FL-assisted Industrial Metaverse.}
\label{fig:fl-assisted-metaverse}
\end{figure}

In the preliminary blueprint of the Industrial Metaverse, managers should be able to control real-world equipment to complete complex tasks, such as robotic arms, in a virtual world using advanced technologies such as VR. As one of the fundamental technologies in the Metaverse, Digital Twins\cite{el2018digital} use sensors scattered all around to collect device data and synchronize to the virtual world in real time. This constant stream of data will be then analyzed by tools such as Machine Learning (ML) to assist managers in decision-making and realizing industrial automation. For example, Optical Character Recognition (OCR) is one of the most widely used ML tools, which helps managers to identify product information and equipment status more efficiently. As a distributed ML paradigm designed for a large number of edge devices and with privacy-preserving capabilities, Federated Learning (FL)\cite{kairouz2021advances} is naturally a promising technique in this context. An illustration of the FL-assisted Industrial Metaverse is shown in Figure \ref{fig:fl-assisted-metaverse}.

However, the fusion of FL and Industrial Metaverse is not a simple combination of technologies. Industrial Metaverse\cite{li2022internet} has extremely demanding requirements for edge intelligence, especially in terms of precision and efficiency\cite{kang2022blockchain}. This poses great challenges from both data and system perspectives under the current industrial network architecture. Specifically, we summarize three main challenges below.

\begin{itemize}
\item[(a)]\textbf{Data Heterogeneity:}
The preference for different classes of data is inherent in every aspect of industrial manufacturing (e.g., among different types of sensors and different functional factories). As a necessary process, manufacturers (e.g., Toyota and Volkswagen) will always prescribe the naming and numbering rules of each component. For example, ``026" and ``034" represent four- and five-cylinder engines and are produced by factories A and B, respectively, and thus A will see more characters ``2" and ``6" than B. Such class biases in the data distribution are called data heterogeneity and have been shown to impair the precision of FL models\cite{zhao2018federated}.

\item[(b)] \textbf{Learning Forgetting:} 
Real-time Digital Twins in Metaverse, combined with In-Memory Computing, enables instant dynamic analysis locally at sensors as streaming data flows in, and allows it to implement real-time industrial systems. However, most sensors have limited memory capacity and cannot store the complete stream of incoming data. For instance, an industrial embedded DDR5 RDIMM has a memory capacity of 16-32GB, but an application can generate gigabits of runtime logs per second. Therefore, old data should be erased from memory to store recent data. Under this paradigm, ML models tend to forget previously learned data knowledge, which is called catastrophic forgetting\cite{french1999catastrophic}.

\item[(c)] \textbf{Limited Bandwidth:}
Wide wireless coverage is often required in open spaces such as mines and blasting sites. The cellular low-power wide-area networks (LPWANs)\cite{chen2022survey}, including NarrowBand IoT, LTE-M, etc., feature long coverage range and low power consumption, and thus become especially favored by the industry. As a price, LPWAN has a low bit rate (e.g., up to 220kbps for NarrowBand IoT and up to 1Mbps for LTE-M\cite{aldahdouh2019survey}), and results in high latency when transmitting ML models and therefore becomes the efficiency bottleneck of FL.
\end{itemize}

Facing the above three challenges, some feasible solutions have already been proposed in the field of FL. However, unfortunately, they are studied separately for a certain challenge, and some are not applicable due to the strict constraints in industrial settings. For the problem of data heterogeneity, some efforts\cite{zhou2016less,jeong2018communication,duan2019astraea} use data augmentation to balance the number of samples in each category. These methods work well on static FL that traverses the local dataset multiple times, but are not suitable for industrial streaming data that can only be seen once. Moreover, the streaming data bring the learning forgetting problem. Related work\cite{rebuffi2017icarl} on continual learning tries to alleviate forgetting by storing historical data, but the storage of a steady stream of industrial data puts great pressure on sensor memory. In terms of system efficiency, compression techniques such as network pruning\cite{hanson1988comparing,srinivas2015data,han2015learning,zhou2016less,wen2016learning,cheng2015exploration}, quantization\cite{gong2014compressing,choi2016towards,courbariaux2015binaryconnect} and sparsification\cite{lin2017deep} have been successful in reducing model size and easing communication burden. However, the pruning-based approaches (e.g., network slimming\cite{liu2017learning}) are data-dependent, which makes their pruned model structure oscillate when dealing with rapidly changing streaming data. The quantization- and sparsification-based approaches have limited compression rates without sacrificing accuracy, and will still burden the small-bandwidth LPWAN network, which requires further compression. 

\begin{figure*}
\centering
\includegraphics[width=.9\textwidth]{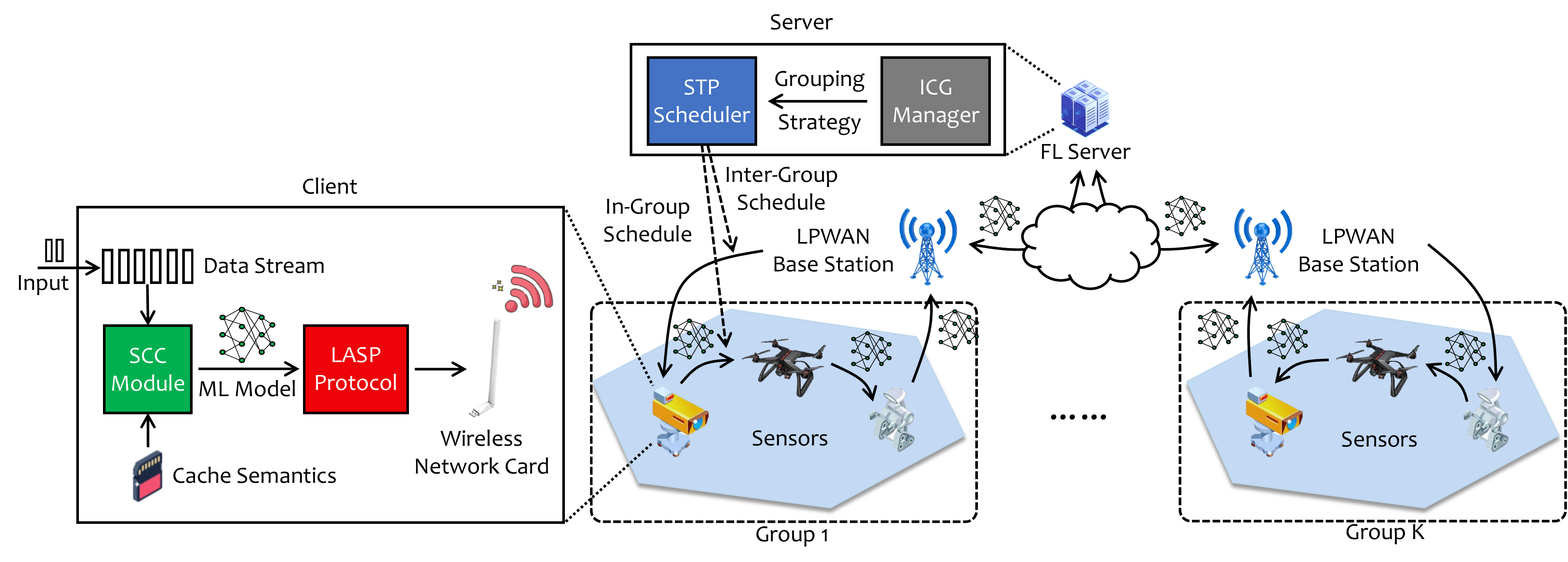}
\caption{The overview of our HFedMS framework.
}
\label{fig:framework}
\end{figure*}

In order to meet the demanding requirements for precision and efficiency of the Industrial Metaverse, FL still lacks practical methods and algorithms to address these three challenges simultaneously. To achieve this goal, this paper presents a novel, high-performance and efficient system, named \NAME. Firstly, \NAME~adopts a new concept of dynamic \textit{Sequential-to-Parallel (STP)} mode to coordinate the training process between devices. Leveraging the robustness of sequential training to data heterogeneity, STP groups devices and encourages devices within each group to train the federated model sequentially, while devices between groups still run in the parallel mode. The grouping function is provided by a fast \textit{Inter-Cluster Grouping (ICG)} algorithm, which outputs device groups with homogeneous data distribution. As training progresses, the number of groups increases, and the system training mode gradually transforms from sequential to fully parallel. Secondly, for the forgetting problem, \NAME~introduces a \textit{Semantic Compression \& Compensation (SCC)} mechanism. The core idea is to store the compressed historical semantics extracted by the representation layer, which will be replayed and compensated and used to calibrate the classifier. Thirdly, a \textit{Layer-wise Alternative Synchronization Protocol (LASP)} is leveraged, which synchronizes the important but lightweight classifier parameters at a higher frequency to reduce the communication burden. These techniques are natively compatible with changing streaming data by design (Figure \ref{fig:framework}). To the best of the authors' knowledge, this is the first work to simultaneously address all three challenges of data heterogeneity, learning forgetting, and limited bandwidth under the strict constraints of the Industrial Metaverse. 

The main contributions are summarized as follows:
\begin{itemize}
\item{We propose a high-performance and efficient FL system \NAME~for Industrial Metaverse, which incorporates a new concept of STP training mode with a fast ICG grouping algorithm, and can effectively resist the heterogeneity of streaming industrial data to improve FL precision.}
\item{We propose SCC, a semantic compression and compensation mechanism that enables historical data to be recorded and stored in a lightweight form on sensors with limited memory. The stored data can be used to calibrate classifier parameters to alleviate learning forgetting.}
\item{We propose LASP, which allocates more communication resources to synchronize important but lightweight classifier parameters. LASP can achieve comparable performance, but with much less traffic, making it well-suited for bandwidth-limited LPWAN networks.}
\item{Extensive experiments are conducted on 368 simulated devices and the streamed non-i.i.d. FEMNIST dataset. \NAME~improves FL accuracy by at least 6.4\% and saves up to 98\% of overall runtime and transfer bytes, proving its superiority in precision and efficiency.}
\end{itemize}

\section{Related Work}
\textbf{Industrial Metaverse and Enabling Technologies.}
The emerging concept of Metaverse\cite{xu2022full} allows for immersive interaction and collaboration between virtual and real worlds. This paradigm is most likely to be adopted by the industrial manufacturing field first, as \textit{Industrial Metaverse}\cite{li2022internet} has a great opportunity to break through the bottlenecks encountered by manufacturers, and is expected to increase productivity and provide better services in the future Internet of Everything. As its four pillar technologies (Figure \ref{fig:fl-assisted-metaverse}), \textit{Digital Twins}\cite{el2018digital,han2022dynamic} collects real-time data from the real-world industrial environment and creates digital avatars of machines and managers in the virtual world. \textit{AI} (especially FL and Blockchain)\cite{kang2022blockchain,jeon2022blockchain,yang2022fusing,chang20226g} learn from these digital data and assist managers in making decisions or executing automated controls, where FL and Blockchain provide privacy and security guarantees respectively. In addition, actions performed in the virtual world will also be synchronized with the real world in real time. \textit{5G/6G High-Speed Communication and Scheduling} and \textit{Mobile Edge Computing}\cite{chang20226g,wang2022mobile,du2022exploring,du2022attention,liu2022slicing4meta} infrastructures guarantee low latency and high reliability for real-time data stream synchronization, aiming to provide users with seamless and immersive interactive experiences.

\textbf{Federated Learning and Heterogeneity.}
Federated Learning (FL), emerged as an advanced distributed ML paradigm that addresses data privacy breaches when working with a large number of smart edge devices, has become an indispensable underlying foundation for Metaverse. A lot of work has been done in FL in terms of efficiency\cite{li2020esync}, privacy\cite{yin2021comprehensive}, security\cite{li2021byzantine,xie2022securing}, etc., but statistical heterogeneity still remains an open problem. In \cite{zhao2018federated}, the authors found that the drop in FL accuracy is caused by the divergence in class distributions between clients (i.e., non-i.i.d.), which can reach a 55\% drop in the worst case. It motivated some efforts\cite{zhao2018federated,jeong2018communication,duan2019astraea} to balance local class distribution by sharing and augmenting data. In \cite{li2020federated}, a proximal penalty term was added to the local loss function to constrain local models. Feature fusion\cite{yao2018two,yao2019towards}, weighted aggregation\cite{yeganeh2020inverse}, client selection \cite{wang2020optimizing}, and adaptive optimizers\cite{reddi2020adaptive} are also promising directions in promoting convergence on non-i.i.d. data. Moreover, the authors in \cite{duan2019astraea,zeng2022heterogeneous,li2022data,li2022fedgs} proposed clustering-based approaches to remove heterogeneity between groups, with best-fit grouping strategies\cite{duan2019astraea,zeng2022heterogeneous} and unique client selection\cite{li2022data,li2022fedgs}.

\textbf{Online and Continual Learning.}
Online learning\cite{hoi2021online} is an ML technology that continuously learns new knowledge from a continuous stream of data (e.g., industrial sensing data), rather than requiring the entire dataset to be prepared in advance as in traditional offline learning. It is a natural trend to combine online learning and FL such as in \cite{han2020adaptive,zhou2019privacy,chen2020asynchronous}, however, the original purpose of online learning did not take into account catastrophic forgetting, where ML models may forget previously learned knowledge in later learning. This phenomenon was first discovered in the study of connectionist networks\cite{french1999catastrophic} and has received increasing attention in recent years\cite{rebuffi2017icarl,pellegrini2020latent,van2020brain,iscen2020memory}. To distinguish it from online learning, continual learning\cite{de2021continual} (also incremental learning) was proposed as a subset of online learning that deals with catastrophic forgetting. For brevity, here we only discuss one of the mainstream directions, that is, data replay. A naive idea is to store the raw HD data and reuse it in later training\cite{rebuffi2017icarl}, but due to the small storage capacity of sensors, doing so will limit the learning ability and put high storage pressure on industrial devices. To this end, lighter and lower-dimensional information should be stored and used for replay. The authors in \cite{pellegrini2020latent} stored the activation map at some intermediate layer, and in \cite{van2020brain} used a standard variational autoencoder to recover past data from stored features. Nonetheless, as training progresses, the stored features may become outdated, but they did not incorporate compensation module to bridge this gap. In \cite{iscen2020memory}, an additional multi-layer perception was used to compensate for outdated features, but this approach increases model complexity and places more burden on devices. 

\begin{algorithm*}
\normalem
\caption{Heterogeneous Federated Learning with Memorable Semantics (\NAME)}\label{alg:hfedms}
Server initializes the global model parameters $\omega^{0}$; \\
Server creates a training scheduler: $\text{STP}\gets\Call{\textbf{Sequential-to-Parallel-Scheduler}}{ }$; \\
\For{each round $r=1,\cdots,R$}{
\% Full synchronization round. \\
\If{every $T$ rounds}{
$\Call{STP.enable\_grouping\_func}{\textsc{\textbf{Inter-Cluster-Grouping}}}$; \\
$\Call{STP.set\_sync\_mode}{\texttt{FULL\_SYNC}}$;  \% \textbf{LASP}: Synchronize the full model parameters. \\
}
\% Calibration round. \\
\Else{
$\Call{STP.enable\_calibration\_func}{\textsc{\textbf{Semantic-Compression-and-Compensation}}}$; \\
$\Call{STP.set\_sync\_mode}{\texttt{PART\_SYNC}}$;  \% \textbf{LASP}: Synchronize only classifier parameters.
}
$\omega^{r}\gets\Call{STP.run}{r}$;
}
\Return $\omega^{r}$;
\end{algorithm*}

In this work, we adopt \cite{li2020federated,yao2018two,yao2019towards,yeganeh2020inverse,reddi2020adaptive} as comparison benchmarks and use ablation experiments to explain why \cite{duan2019astraea,zeng2022heterogeneous} is not applicable for our highly dynamic case. Then, we propose a novel replay mechanism that stores lightweight semantic features and leverages a \textit{low-footprint} compensation function to calibrate classifier parameters. As another important feature, our approach \textit{reduces considerable traffic without any loss of information}. These features are invaluable and superior to commonly used compression methods that can compromise model accuracy (e.g., sparsification\cite{lin2017deep} and quantization\cite{seide20141}) or slow down training (e.g., low-rank decomposition\cite{liu2015l_}).
\section{Heterogeneous Federated Learning with Memorable Data Semantics}
In this section, we first introduce the overall blueprint of our \NAME~system. Then, the concepts and design details of the supporting techniques are described respectively.

\subsection{The Overall System}
The system overview of \NAME~is illustrated in Figure \ref{fig:framework}. The basic hardware facilities include various types of smart sensors, LPWAN base stations, and an FL server in the cloud. 
Smart sensors are advanced sensors capable of sensing, storing, processing, and transmitting streaming sensor data from surrounding objects.
As factory automation advances, more sophisticated sensors are driving industrial manufacturing (e.g. smart cameras, drones, and robots) to work in smarter ways. For example, the integration of FPGA, ASIC (TPU, NPU, VPU, etc.) circuit chips and AI software enables sensors to have sufficient computational resources and storage capacity to run ML algorithms, which makes them intelligent. 

These smart sensors cache historical data semantics and run the SCC module to compensate and calibrate local model training, then the local model parameters are synchronized following the LASP protocol. They also establish wireless communication channels with nearby LPWAN base stations, while LPWAN gateways connect to the cloud FL server over the backbone network.
LPWAN, typically NarrowBand IoT, LTE-M, etc., realizes high reliability, long-distance coverage (up to more than 50km) of wireless transmission with low power consumption, and has an excellent ability to penetrate obstacles, making it one of the most promising technologies for smart city and manufacturing. Moreover, communication between smart sensors can also be achieved through LPWAN base station relays.

The cloud FL server can be a general application server that runs basic FL functional modules such as device selection, model distribution, aggregation, and updating, as well as the ICG manager and the STP scheduler proposed in this work. The ICG manager decides the grouping strategy of sensors, where the strategy will serve as the basis for STP to schedule the training topology between the devices.

Here we briefly summarize the workflow of \NAME. First, We introduce the LASP synchronization protocol. LASP alternately performs two synchronization phases, that is, full synchronization and calibration. To distinguish these two phases, we define $T$ as the round interval at which full synchronization is performed. In other words, $T-1$ calibration rounds should be performed after every full synchronization round. 

At the beginning of training, the FL server initializes global model parameters and creates the STP scheduler to coordinate training between devices. Then in each full synchronization round, the STP scheduler calls the ICG module to group sensor devices, and some of the groups will be selected to participate in this round and subsequent $T-1$ calibration rounds of training. In addition, according to the LASP protocol, STP runs in \texttt{FULL\_SYNC} mode and the complete model parameters will be synchronized between devices. 

Instead, in a calibration round, ICG is off (keep using previously assigned groups) but SCC is on. SCC uses the compressed semantics of cached historical data to calibrate classifier parameters, in order to avoid it overfitting to recent data but forgetting past knowledge, with a view to solving the learning forgetting problem. Moreover, LASP sets STP to run in \texttt{PART\_SYNC} mode at this time, in which only the parameters of classifiers could be synchronized. 

The overall training process is controlled by the STP scheduler, which decides how and in which order the devices within and between groups deliver their local models. These steps will repeat for $R$ times to obtain a well-trained FL model.
We summarize the relationship of the four proposed techniques (STP, ICG, SCC, and LASP) in Algorithm \ref{alg:hfedms}, and their design details will be elaborated in subsequent subsections.

\subsection{STP: Sequential-to-Parallel Training} 

\begin{figure*}
\centering
\includegraphics[width=0.75\textwidth]{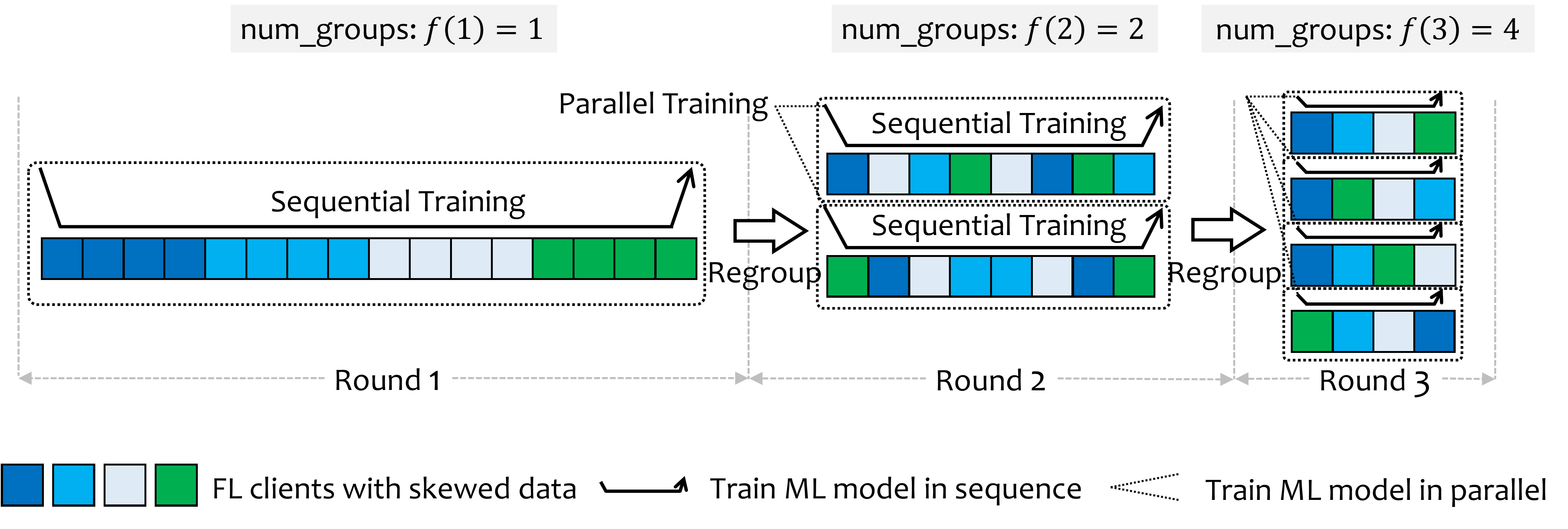}
\caption{A simplified example of STP workflow. The model within each group is trained in sequence, while the models between groups are trained in parallel. At round $r$, the number of groups is increased according to $f(\cdot)$, and clients are shuffled after each regroup.}
\label{fig:stp}
\end{figure*}

The problem of data heterogeneity has been a hot topic in FL for a long time, and unfortunately, in industrial manufacturing, heterogeneous data can be found everywhere. This problem arises from clients' preferences for different objects, actions, and properties, its impact is reflected in skewed local data distributions and can impair the convergence of many common parallel algorithms, notably FedAvg\cite{zhao2018federated}. Conversely, some work\cite{duan2019astraea} has shown that training clients sequentially is less affected by data heterogeneity, that is, clients train local models received from their predecessor clients and deliver the trained local models to their successor clients. In some cases, such as the local dataset is properly shuffled and traversed for only one epoch, this sequential training mode is theoretically equivalent to the centralized SGD algorithm, making it inherently robust to data heterogeneity.

Following this idea, Duan et al.\cite{duan2019astraea} proposed an intuitive approach, in which they assign clients to multiple groups and force the overall data distributions between the groups to be homogeneous. Then, faced with clients that still have heterogeneous data in the group, they adopt the sequential training mode to jointly train a model. On the other hand, parallel algorithms (e.g., FedAvg\cite{mcmahan2017communication}) can be used between groups as their data becomes homogeneous. Their approach obtains noticeable gains in model performance on non-i.i.d. data. However, they only considered a static environment where local datasets do not change, making it unsuitable for highly dynamic industrial practices such as streaming data.

In this section, we present a more dynamic \textit{Sequential-to-Parallel (STP)} training mode to accommodate the ever-changing streaming data. An example of STP workflow is shown in Figure \ref{fig:stp}. In each full synchronization round $r(\forall r\le R, r \bmod T=0)$, STP reassigns clients into $M=f(r)$ groups and shuffles their order, where $f(\cdot)$ is a growing function that controls the number of groups. Similarly, the clients within each group are trained in sequence, and the clients between groups are trained in parallel. As rounds progress, clients are split into more groups and each group contains fewer clients. This means that the FL process is gradually transforming from (fully) sequential to (fully) parallel training.

The choice of function $f(\cdot)$ has a non-negligible effect on the performance of STP. Here we recommend three typical growth functions: linear (smooth), logarithmic (first fast and then slow), and exponential (first slow and then fast), as follows:
\begin{align}
\nonumber
&\mathrm{Linear:} & f(r)=&\beta\left\lfloor\alpha (\frac{r}{T}-1)+1\right\rfloor, \\
\nonumber
&\mathrm{Logarithmic:} & f(r)=&\beta\left\lfloor\alpha\ln (\frac{r}{T})+1\right\rfloor, \\
\nonumber
&\mathrm{Exponential:} & f(r)=&\beta\lfloor(1+\alpha)^{\frac{r}{T}-1}\rfloor,
\end{align}
where $T$ is the round interval, $\alpha$ is a real number controlling the growth rate, and $\beta$ is an integer controlling the initial number of groups and the growth span. In the experiments, we will explore the best growth function and recommend the corresponding $\alpha,\beta$ settings.

The detailed implementation is given in Algorithm \ref{alg:stp}. To simplify the description, we only consider the case where \texttt{CALIBRATION\_ENABLED=FALSE} and the sync mode is \texttt{FULL\_SYNC} (i.e., a full synchronization round), while other cases will be explained in subsequent subsections. In each full synchronization round $r (\forall r\le R, r \bmod T=0)$, STP reassigns clients to $M=f(r)$ groups via ICG and selects a small proportion of $\kappa$ groups at random to participate in the next $T$ rounds of training. Then, the selected groups perform the following operations in parallel: The first client of each group pulls the latest global model from the FL server and delivers it to the next client in order after local training, until the last client pushes the sequentially trained model to the FL server. When receiving model parameters from preorder clients, these clients traverse the batch of streaming data just captured for one epoch, and train the model using typical optimizers such as SGD and Adam. Finally, the models of all groups are aggregated at the FL server to update the global model.

\begin{algorithm}[t]
\normalem
\caption{Sequential-to-Parallel Scheduler (STP)}
\label{alg:stp}
\KwIn{The current round $r$, the group growth function $f(\cdot)$, the group sample ratio $\kappa$, the learning rate $\eta$, the batch size $b$.}
\KwOut{The updated federated model $\omega^{r}$.}
\If{{\rm\texttt{GROUPING\_ENABLED}}}{
Increase the number of groups $M=f(r)$; \\
$\mathcal{G}\gets\Call{\textbf{Inter-Cluster-Grouping}}{M}$; \\
Select a subset of groups $\tilde{\mathcal{G}}\subset\mathcal{G}$ with proportion $\kappa$;
}
\For{each group $\mathcal{G}_{m}$ {\rm\textbf{in}} $\tilde{\mathcal{G}}$ in parallel}{
\If{\rm\texttt{FULL\_SYNC}}{
Let $\omega$ denote the parameters of the full model;
}
\ElseIf{\rm\texttt{PART\_SYNC}}{
Let $\omega$ denote the parameters of the classifier;
}
The first client $\mathcal{C}_m^1$ in $\mathcal{G}_{m}$ pulls $\omega_{m}^{1}\gets\omega^{r-1}$; \\
\For{each client $\mathcal{C}_{m}^{k}$ {\rm\textbf{in}} $\mathcal{G}_m$ in sequence}{
Capture a batch of streaming data $\mathcal{D}_{m}^{k}$; \\
\If{\rm\texttt{CALIBRATION\_ENABLED}}{
Construct the calibration dataset $\tilde{\mathcal{D}}_{m}^{k}\gets$\textsc{\textbf{Semantic-Compression-and-Compensation}}$(\mathcal{D}_{m}^{k})$; \\
}
Train $\omega_m^k$ for one epoch using SGD or Adam with learning rate $\eta$ and mini-batch size $b$; \\
Send $\omega_{m}^{k}$ to next client $\mathcal{C}_m^{k+1}$, $\omega_{m}^{k+1}\gets\omega_{m}^{k}$;
}
The last client $\mathcal{C}_m^{|\mathcal{G}_m|}$ uploads $\omega_m^{|\mathcal{G}_m|}$;
}
Server aggregates $\omega^{r}\gets\frac{1}{\kappa M}\sum_{\forall\mathcal{G}_m\in\tilde{\mathcal{G}}}{(\omega_m^{|\mathcal{G}_m|})}$; \\
Server scatters $\omega^{r}$ to clients if in \texttt{FULL\_SYNC} mode; \\
\Return $\omega^{r}$;
\end{algorithm}

This design of STP has several advantages:
\begin{enumerate}
\item[(a)] STP solves the problem of data heterogeneity because the negative effects of within-group heterogeneity are attenuated by sequential training and between-group heterogeneity is removed by the grouping mechanism.
\item[(b)] Clients are regrouped every $T$ rounds, which makes STP dynamic and adaptable to local changing streaming data.
\item[(c)] The growing number of groups improves system parallelism, which can speed up training when FL is close to convergence. Moreover, this design prevents forgetting caused by a long ``chain of clients'' in a group. That is, the model may forget the data of previous clients and overfit the data of subsequent clients.
\item[(d)] STP shuffles clients after each grouping, preventing models from learning interfering information, such as the order of clients.
\end{enumerate}

\subsection{ICG: Inter-Cluster Grouping}
As described in the previous section, STP should regroup clients at each full synchronization round, and these groups are required to have homogeneous data distributions. These strict requirements place higher demands on the quality and runtime of the solution, making it a difficult problem to solve. In this subsection, we first formulate the mathematical model of the client grouping problem, and then introduce our fast ICG algorithm to solve it.

Consider an $\mathcal{F}$-class FL classification task involving $K$ clients that should be assigned to $M$ groups. These clients (sensors) are capturing streaming data all the time, and they batch the enqueued data to train local models in every round. When grouping is to be performed, the client reports a summary of the statistical distribution of the current batch of data to the ICG manager. This summary will serve as the basis for ICG to group clients. 

Our goal is to find a grouping strategy $\mathbf{x}\in\mathbb{I}^{M\times K}$ in the 0-1 space $\mathbb{I}=\{0,1\}$ to minimize the difference in class distributions of all groups, where $\mathbf{x}_m^k=1$ represents the client $k$ is assigned to the group $m$, $\mathcal{V}\in(\mathbb{Z^+})^{\mathcal{F}\times K}$ is the class distribution matrix composed of $\mathcal{F}$-dimensional class distribution vectors of $K$ clients, $\mathcal{V}_m\in(\mathbb{Z}^+)^{\mathcal{F}\times 1}$ represents the overall class distribution of group $m$, and $\left<\cdot,\cdot\right>$ represents the distance between two class distributions. The problem can be formulated as follows:
\begin{align}
\underset{\mathbf{x}}{\mathrm{minimize}}\qquad & z=\sum_{m_1=1}^{M-1}\sum_{m_2=m_1+1}^{M}<\mathcal{V}_{m_1},\mathcal{V}_{m_2}>, \label{eq:objective}\\
\mathrm{s.t.}\qquad & M=f(r), \label{eq:group-number}\\
& \sum_{k=1}^{K}\mathbf{x}_m^k\le \left\lceil \frac{K}{M} \right\rceil \quad \forall m=1,\cdots,M, \label{eq:group-capacity}\\
& \sum_{m=1}^{M}{\mathbf{x}_m^k}=1 \qquad\quad \forall k=1,\cdots,K, \label{eq:client-conflict}\\
& \mathcal{V}_m=\sum_{k=1}^{K}{\mathbf{x}_m^k\mathcal{V}^k}\quad \forall m=1,\cdots,M, \label{eq:overall-dist}\\
& \mathbf{x}_m^k\in\{0,1\}, ~k\in[1,K], ~m\in[1,M].
\label{eq:variable-constraint}
\end{align}
Constraint \eqref{eq:group-capacity} simplifies the problem and ensures that the groups have similar or equal size $\left\lceil \frac{K}{M} \right\rceil$. Constraint \eqref{eq:client-conflict} ensures that each client can only be assigned to one group at a time. The overall class distribution $\mathcal{V}_m$ of the group $m$ is defined by Eq. \eqref{eq:overall-dist}, where $\mathcal{V}^k\in\mathcal{V}$ is the class distribution vector of client $k$. This problem can reduce to an NP-hard bin packing problem, thus it is almost impossible to find the optimal solution within a polynomial time.

To solve this problem, we simplify the original problem to obtain a constrained clustering problem. Consider a constrained clustering problem with $K$ points and $L$ clusters, where all clusters are of the same size $\lfloor K/L\rfloor$. We make the following assumptions.

\vspace{1mm}
\begin{assumption}
\label{icg-assumption}
(a) $K$ is divisible by $L$;
(b) Take any point $\mathcal{V}^m_l$ from cluster $l$, the squared $l_2$-norm distance $\|\mathcal{V}^m_l-C_l\|_2^2$ between the point $\mathcal{V}^m_l$ and its cluster centroid $C_l$ is bounded by $\sigma_l^2$.
(c) Take one point $\mathcal{V}^m_l$ from each of $L$ clusters at random, the sum of deviations of each point from its cluster centroid $\epsilon^m=\sum_{l=1}^{L}(\mathcal{V}^m_l-C_l)$ satisfies $\mathbb{E}[\epsilon^m]=0$.
\end{assumption}

\vspace{1mm}
\begin{definition}[Group Centroid]\label{def:group-centroid}
Given $L$ clusters of equal size, let group $m$ be constructed from one point randomly sampled from each cluster $\{\mathcal{V}^m_1,\cdots,\mathcal{V}^m_L\}$. Then, the centroid of  group $m$ is defined as $C^m=\frac{1}{L}\sum_{l=1}^{L}\mathcal{V}^m_l$.
\end{definition}

\vspace{1mm}
\begin{proposition}\label{prop:centroid}
Let Assumption \ref{icg-assumption} hold, suppose the centroid of cluster $l$ is $C_l=\frac{L}{K}\sum_{i=1}^{K/L}{\mathcal{V}_l^i}$ and the global centroid is $C_\mathrm{global}=\frac{1}{L}\sum_{l=1}^{L}{C_l}$. We have:
\begin{enumerate}
\item The group and global centroids are expected to coincide, $\mathbb{E}[C^m]=C_\mathrm{global}$.
\item The error $\|C^m-C_\mathrm{global}\|_2^2$ between the group and global centroids is bounded by $\frac{1}{L^2}\sum_{l=1}^{L}{\sigma_l^2}$.
\end{enumerate}
\end{proposition}
\vspace{1mm}

\begin{algorithm}[t]
\normalem
\caption{Inter-Cluster-Grouping (ICG)}\label{alg:icg}
\KwIn{The set of clients $\mathcal{C}$, the number of clients $K$, the reported data distribution $\mathcal{V}^k$ for each client $k$, the number of groups $M$.}
\KwOut{The grouping strategy $\mathcal{G}$.}
Sample $L\cdot\lfloor\frac{K}{L}\rfloor$ clients from $\mathcal{C}$ at random to satisfy Assumption \ref{icg-assumption}, where $L=\lfloor\frac{K}{M}\rfloor$; \\
\Repeat{$C_l$ converges}{
\textsc{Cluster Assignment:} Fix the cluster centroid $C_l$ and optimize $\mathbf{y}$ in Eq. \eqref{eq:ccp-objective} to Eq. \eqref{eq:ccp-value}; \\
\textsc{Cluster Update:} Fix $\mathbf{y}$ and update the cluster centroid $C_l$ as follows, $$C_l\gets\frac{\sum_{k=1}^{K}{\mathbf{y}_l^k\mathcal{V}^k}}{\sum_{k=1}^{K}{\mathbf{y}_l^k}} \quad \forall l=1,\cdots,L;$$}
\For{$m=1,\cdots,M$}{
\textsc{Group Assignment:} Sample one client from each cluster at random without replacement to construct group $\mathcal{G}_m$; \\
Shuffle the order of clients in group $\mathcal{G}_m$;
}
\Return $\mathcal{G}=\{\mathcal{G}_1,\cdots,\mathcal{G}_{M}\}$;
\end{algorithm}

Please refer to Appendix \ref{proof:prop1} for the proof. Proposition \ref{prop:centroid} indicates that there exists a grouping strategy $\tilde{\mathbf{x}}$ and $\mathcal{V}_{m_1}=\sum_{k=1}^{K}\tilde{\mathbf{x}}_{m_1}^k\mathcal{V}^k=LC^{m_1}$, $\mathcal{V}_{m_2}=\sum_{k=1}^{K}\tilde{\mathbf{x}}_{m_2}^k\mathcal{V}^k=LC^{m_2}$ ($\forall m_1\ne m_2$), so that the objective in Eq. \eqref{eq:objective} turns to $z=\sum_{m_1\ne m_2}L<C^{m_1},C^{m_2}>$  and the expectation value reaches 0. This motivates us to use the constrained clustering model to solve $\tilde{\mathbf{x}}$. Therefore, we consider the constrained clustering problem defined as follows,
\begin{align}
\underset{\mathbf{y}}{\mathrm{minimize}}\qquad & \sum_{k=1}^{K}\sum_{l=1}^{L}\mathbf{y}^k_l\cdot\left(\frac{1}{2}\|\mathcal{V}^k-C_l\|_2^2\right), \label{eq:ccp-objective}\\
\mathrm{s.t.}\qquad & \sum_{k=1}^{K}\mathbf{y}_l^k=\frac{K}{L} \qquad \forall l=1,\cdots,L, \label{eq:ccp-least}\\
& \sum_{l=1}^{L}\mathbf{y}_l^k=1 \qquad\quad \forall k=1,\cdots,K, \label{eq:ccp-client}\\
& \mathbf{y}_l^k\in\{0,1\}, ~k\in[1,K], ~l\in[1,L], \label{eq:ccp-value}
\end{align}
where $\mathbf{y}\in\mathbb{I}^{L\times K}$ is a selector variable, $\mathbf{y}_l^k=1$ means that client $k$ is assigned to cluster $l$ while 0 means not, $C_l$ represents the centroid of cluster $l$. Eq. \eqref{eq:ccp-objective} is the standard clustering objective, which aims to assign $K$ clients to $L$ clusters so that the sum of the squared $l_2$-norm distance between the class distribution vector $\mathcal{V}^k$ and its nearest cluster centroid $C_l$ is minimized. Constraint \eqref{eq:ccp-least} ensures that each cluster has the same size $\frac{K}{L}$. Constraint \eqref{eq:ccp-client} ensures that each client can only be assigned to one cluster at a time. In this simplified problem, Constraint \eqref{eq:client-conflict} is relaxed to $\sum_{m=1}^{M}{\mathbf{x}_m^k}\le 1$ to satisfy the assumption that $K/L$ is divisible.

The problem Eq. \eqref{eq:ccp-objective}-\eqref{eq:ccp-value} can be further modeled as a minimum cost flow (MCF) problem and solved by network simplex algorithms \cite{bradley2000constrained}. Then we can alternately perform cluster assignment and cluster update to optimize $\mathbf{y}_l^k$ and $C_l (\forall k,l)$, respectively. Finally, we construct $M$ groups and shuffle their clients. Each group consists of one client randomly sampled from each cluster without replacement, so that their group centroids are expected to coincide with the global centroid. The pseudo-code is given in Algorithm \ref{alg:icg}. The complexity of ICG is $\mathcal{O}(\frac{K^6\mathcal{F}\tau}{M^2}\log{Kd})$, where $d=\max\{\sigma_l^2 | \forall l\in[1,L]\}$, and $\tau$ is the number of  iterations. In preliminary experiments, ICG runs fast and can complete group assignment within 100 milliseconds, with $K=368,M=52,\mathcal{F}=62$ and $\tau=10$.

\subsection{SCC: Semantic Compression \& Compensation}
Traditional ML has a fixed dataset that can be traversed over many epochs, but this is not true in industrial applications because sensors are constantly collecting data and there is not enough memory to store this long-term streaming data. As a result, training samples should be discarded after being used once, and the ML model will gradually forget the information of historical samples, which is called learning forgetting (or catastrophic forgetting)\cite{french1999catastrophic}.

\begin{figure*}
\centering
\includegraphics[width=.7\textwidth]{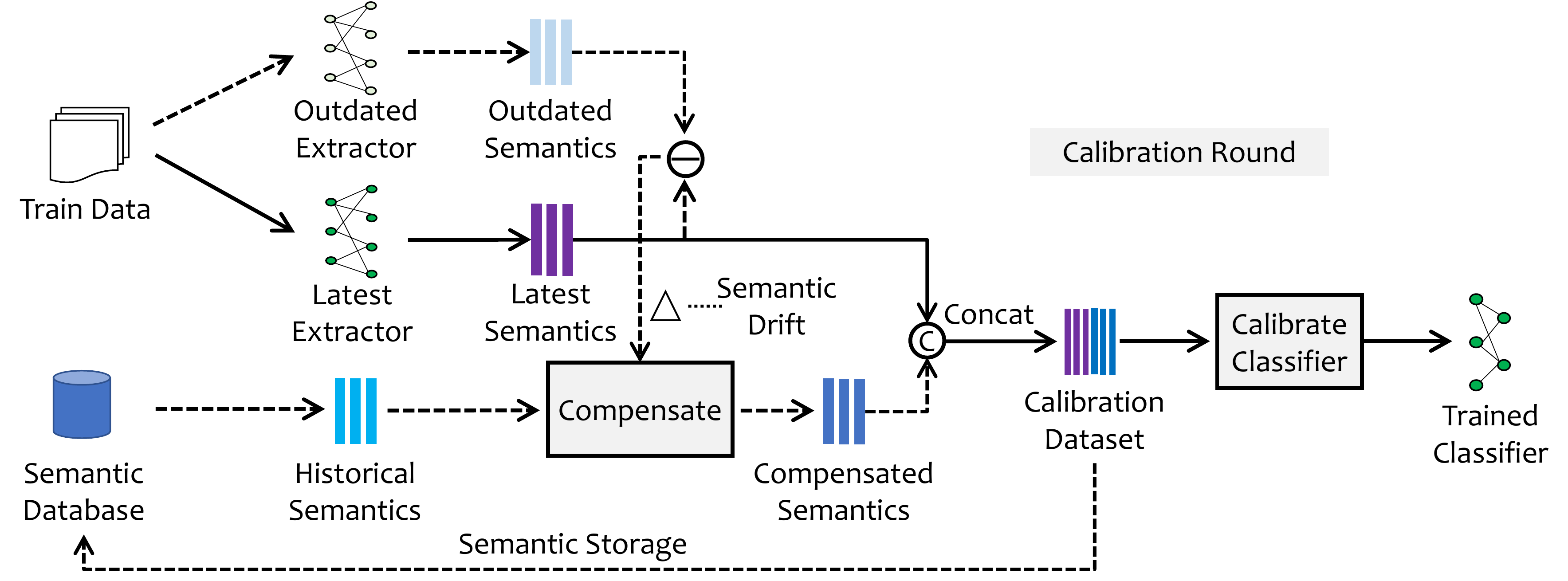}
\caption{Illustration of SCC workflow. In a calibration round, the client backs up the outdated extractor to estimate the semantic drift and uses it to compensate for historical semantics. Then, the latest semantics and historical semantics are fused to train the classifier.}
\label{fig:scc}
\end{figure*}

In this subsection, we propose a semantic-based compression and compensation approach, named SCC, to make long-term historical data recordable and calibrate classifier parameters to alleviate forgetting. As illustrated in Figure \ref{fig:scc}, SCC stores the semantic features extracted by the feature representation layer and adaptively compensates for outdated semantics. The semantic features are low-dimensional feature embeddings of the raw input data (e.g., vectors of dimension 100), and their number of bytes stored is much less than the raw data (e.g., HD video streams), making the long-term data streams lightweight and recordable. Then these cached data semantics can be used to extend the current data semantics and calibrate the training of classifier parameters.

Let us formally describe the workflow of SCC. Let $\theta$ be the parameters of the classifier, $\phi$ be the parameters of the feature extractor, $\omega=\{\phi,\theta\}$ be the parameters of the full model, and $g,h,f$ be the forward functions, respectively. The classification task can be represented by $$f(x, \omega)=g(z, \theta), z=h(x, \phi),$$ where $z$ is the semantic feature in a low-dimensional latent space of data point $x$.

Suppose that at the $r$-th round ($r\bmod T=0$, i.e., a full synchronization round), the group $\mathcal{G}_m$ is selected and the global parameters of the extractor after running STP is $\phi^r$. Then, in the subsequent $l$-th calibration round ($l<T$), for any client $\mathcal{C}_m^k$ in the group $\mathcal{G}_m$, let $\mathcal{D}_m^k(r+l)$ denote the batch of streaming data available in current round, and $x_i\in\mathcal{D}_m^k(r+l)$ denote the $i$-th training sample in $\mathcal{D}_m^k(r+l)$. Furthermore, we define $\tilde{r}$ ($\tilde{r}<r$, also a full synchronization round) as the previous round in which the client $\mathcal{C}_m^k$ was also selected, and $\phi^{\tilde{r}}$ as the corresponding global parameters at that round.

As shown in Figure \ref{fig:scc}, a semantic database runs on the client to help store semantic features and is represented by $\mathcal{Z}_m^k$. Then, a naive idea emerged, that is to directly fuse the current data semantics $\mathcal{S}_m^k(r+l)=h(\mathcal{D}_m^k(r+l),\phi_m^{k, r+l-1})$ with the historical semantics $\mathcal{Z}_m^k$, and use the fused one to train the classifier $\theta_m^{k, r+l-1}$, where $\phi_m^{k, r+l-1}=\phi^{r},\forall l=1,\cdots,T-1$, since the feature extractor $\phi$ is frozen in calibration rounds. 

However, the feature extractor will be updated every $T$ rounds, which means that we are fusing the historical semantics extracted by the outdated extractor with the latest semantics extracted by the current extractor. This mismatch can lead to confusion. More formally, suppose that at the $\tilde{r}$-th round, the semantics $\mathcal{Z}_m^k$ is generated using the dataset $\mathcal{D}_m^k(\tilde{r})$ and the parameters $\phi^{\tilde{r}}$, then in the $(r+l)$-th round ($r-\tilde{r}\ge T$), the historical semantics $\mathcal{Z}_m^k$ and their correct values ${\mathcal{Z}_{m}^{k}}^{*}$ should be
\begin{align}
\nonumber
&\mathcal{Z}_m^k=h(\mathcal{D}_m^k(\tilde{r}),\phi^{\tilde{r}}), \\
\nonumber
&{\mathcal{Z}_m^k}^*=h(\mathcal{D}_m^k(\tilde{r}),\phi^r), \\
\nonumber
\text{s.t.~}& r-\tilde{r}\ge T \text{~and~} \phi^{\tilde{r}}\ne\phi^r.
\end{align}
Clearly, we have $\mathcal{Z}_m^k\ne{\mathcal{Z}_m^k}^*$ and there exists a gap between them. For ease of understanding, we consider a feature extractor $h(x,\phi)$ composed of fully connected layers or convolutional layers (excluding nonlinear layers), where $\phi=\{\phi_1, \phi_2, \cdots, \phi_L\}$, $\phi_i$ denotes the parameters of the $i$-th layer in $\phi$, and $L=|\phi|$ denotes the number of layers in $\phi$. The forward function $h(x,\phi)$ can be written as follows,
\begin{equation}
\nonumber
h(x,\phi)=x\cdot\phi_1\cdot\phi_2\cdots\phi_L.
\end{equation}
To simplify the analysis, we define $\Phi=\prod_{i=1}^{L}{\phi_i}$ and have $h(x,\phi)=x\cdot\Phi$. Then, the gap ${\mathcal{Z}_m^k}^*-\mathcal{Z}_m^k$ can be given by
\begin{align}
\nonumber
{\mathcal{Z}_m^k}^*-\mathcal{Z}_m^k&=h(\mathcal{D}_m^k(\tilde{r}),\phi^r)-h(\mathcal{D}_m^k(\tilde{r}),\phi^{\tilde{r}}) \\
\nonumber
&=\mathcal{D}_m^k(\tilde{r})\cdot\Phi^r-\mathcal{D}_m^k(\tilde{r})\cdot\Phi^{\tilde{r}} \\
&=\mathcal{D}_m^k(\tilde{r})\cdot(\Phi^r-\Phi^{\tilde{r}}) \label{eq:scc-gap-l3}.
\end{align}
Nonetheless, Eq. \eqref{eq:scc-gap-l3} still cannot be used to compensate for $\mathcal{Z}_m^k$ because the historical data $\mathcal{D}_m^k(\tilde{r})$ is unknown. Therefore, we define the semantic drift $\Delta$ as the expectation of ${\mathcal{Z}_m^k}^*-\mathcal{Z}_m^k$ and make Assumption \ref{scc-assumption} to bridge the relationship between the historical data $\mathcal{D}_m^k(\tilde{r})$ and the current data $\mathcal{D}_m^k(r+l)$.

\vspace{1mm}
\begin{assumption}
\label{scc-assumption}
For any client $\mathcal{C}_m^k$, the data batches used in any two rounds $r_1,r_2$ have similar distributions in expectation, that is, 
$\mathbb{E}_{x_i}[\mathcal{D}_m^k(r_1)]=\mathbb{E}_{x_i}[\mathcal{D}_m^k(r_2)]+\Xi,\forall r_1\ne r_2$ where $\Xi$ is a tensor of deviations with $|\Xi|\rightarrow 0$ and $\mathbb{E}_{r}[\Xi]=\boldsymbol{0}$.
\end{assumption}
\vspace{1mm}

This assumption is reasonable because the statistical distribution of industrial data is often stable, and dynamic personalization is not the focus of this work. Let Assumption \ref{scc-assumption} hold, the semantic drift $\Delta$ can be written as follows,
\begin{align}
\nonumber
\Delta&=\mathbb{E}_{x_i}[{\mathcal{Z}_m^k}^*-\mathcal{Z}_m^k]=\mathbb{E}_{x_i}[\mathcal{D}_m^k(\tilde{r})\cdot(\Phi^r-\Phi^{\tilde{r}})] \\
\nonumber
&=\mathbb{E}_{x_i}[\mathcal{D}_m^k(\tilde{r})]\cdot(\Phi^r-\Phi^{\tilde{r}}) \\
\nonumber
&=\mathbb{E}_{x_i}[\mathcal{D}_m^k(r+l)]\cdot(\Phi^r-\Phi^{\tilde{r}})+\Xi\cdot(\Phi^r-\Phi^{\tilde{r}}),
\end{align}
then we can use $\Delta$ to compensate for the stored historical semantics $\mathcal{Z}_m^k$,
\begin{align}
{\mathcal{Z}_m^k}^{*}&=\mathcal{Z}_m^k+\Delta \label{eq:scc-compensate} \\
\nonumber
&=\mathcal{Z}_m^k+\mathbb{E}_{x_i}[\mathcal{D}_m^k(r+l)]\cdot(\Phi^r-\Phi^{\tilde{r}})+o(\Phi^r-\Phi^{\tilde{r}}),
\end{align}
where $o(\Phi^r-\Phi^{\tilde{r}})$ is a negligible term close to zero.

\vspace{1mm}
\textit{(a) Semantic Compensation.} 
Following the design of Eq. \eqref{eq:scc-compensate}, we backup the outdated parameters $\phi^{\tilde{r}}$ on the client to assist with semantic compensation. It should be noted that not all historical parameters will be backed up. To avoid ambiguity, we use $r_{\mathrm{bak}}$ instead of $\tilde{r}$ to represent the previous (full synchronization) round when the current client was selected, and use $r$ instead of $r+l$ to represent the current (calibration) round. A standard semantic compensation process can be implemented by
\begin{gather}
\nonumber
\mu^r\gets\mathbb{E}_{x_i}[h(x_i,\phi^{r-1})|(x_i, y_i)\in\mathcal{D}_m^k(r)], \\
\nonumber
\mu^{r_{\mathrm{bak}}}\gets\mathbb{E}_{x_i}[h(x_i,\phi^{r_{\mathrm{bak}}})|(x_i,y_i)\in\mathcal{D}_m^k(r)], \\
\nonumber
{\mathcal{Z}_m^k}^{*}\gets\mathcal{Z}_m^k+(\underbrace{\mu^r - \mu^{r_{\mathrm{bak}}}}_{\Delta}).
\end{gather}

Furthermore, for a classification task, we can estimate the semantic drift $\Delta_c$ of different classes $c\in\mathcal{F}$ separately for more accurate compensation,
\begin{gather}
\mu_c^r\gets\mathbb{E}_{x_i}[h(x_i,\phi^{r-1})|(x_i,y_i)\in\mathcal{D}_m^k(r), y_i=c], \label{eq:scc-compensation-1}\\
\mu_c^{r_{\mathrm{bak}}}\gets\mathbb{E}_{x_i}[h(x_i,\phi^{r_{\mathrm{bak}}})|(x_i,y_i)\in\mathcal{D}_m^k(r), y_i=c], \\
{\mathcal{Z}_m^k}^{*}(c)\gets\mathcal{Z}_m^k(c)+(\underbrace{\mu_c^r - \mu_c^{r_{\mathrm{bak}}}}_{\Delta_c}), \forall c\in\mathcal{F}. \label{eq:scc-compensation-3}
\end{gather}

\vspace{1mm}
\begin{proposition}
\label{prop:semantic-drift}
Given Assumption \ref{scc-assumption} and any two adjacent calibration rounds $r_1$ and $r_2$, there exists $r=kT, k\in\mathbb{Z}^+$ such that $r<r_1<r_2<r+T$ and $\Phi^r=\Phi^{{r_1}-1}=\Phi^{{r_2}-1}$, and then we have the difference of semantic drifts $\Delta^{r_1}-\Delta^{r_2}$,
\begin{gather}
\nonumber
\mathbb{E}_r[\Delta^{r_1}-\Delta^{r_2}]=\boldsymbol{0}, \\
\nonumber
\mathrm{Cov}(\Delta^{r_1}-\Delta^{r_2})=(\Phi^r-\Phi^{\tilde{r}})^T\cdot\mathbb{E}_r[\Xi^T\cdot\Xi]\cdot(\Phi^r-\Phi^{\tilde{r}})\rightarrow\boldsymbol{0},
\end{gather}
where 
\begin{align}
\nonumber
\Delta^{r_1}&=\mathbb{E}_{x_i}[\mathcal{D}_m^k(r_1)]\cdot(\Phi^{{r_1}-1}-\Phi^{\tilde{r}}), \\
\nonumber
\Delta^{r_2}&=\mathbb{E}_{x_i}[\mathcal{D}_m^k(r_2)]\cdot(\Phi^{{r_2}-1}-\Phi^{\tilde{r}}).
\end{align}
\end{proposition}
\vspace{1mm}

\begin{algorithm}[t]
\normalem
\caption{Semantic-Compression-and-Compensa-tion (SCC)}\label{alg:lasp}
\KwIn{Client $\mathcal{C}_m^k$, current round $r$, current dataset $\mathcal{D}_m^k$, interval $T$, storage capacity $Q$.}
\KwOut{Calibration dataset $\tilde{\mathcal{D}}_m^k$.}
Freeze the local feature extractor $\phi_m^{k,r-1}$; \\
Extract the semantics of the current dataset
$\mathcal{S}_m^k(r)=h(\mathcal{D}_m^k(r),\phi_m^{k,r-1})$; \\
\% Compensation occurs in the first calibration round. \\
\If{$r\bmod T=1$}{
\textsc{Semantic Compensation:} Compensate the historical semantics by Eqs. \eqref{eq:scc-compensation-1}-\eqref{eq:scc-compensation-3} and obtain ${\mathcal{Z}_m^k}^*$; \\
}
\Else{Reuse the compensated semantics ${\mathcal{Z}_m^k}^*$;}
\textsc{Calibration}: Construct the calibration dataset $\tilde{\mathcal{D}}_m^k$ by Eq. \eqref{eq:scc-calibration-dataset}; \\
\% Update memory in the last calibration round \\
\If {$(r+1) \bmod T=0$}{
\textsc{Semantic Storage}: Store the $Q$ semantics of the smallest value of $\delta_i$ in $\mathcal{Q}$, as in Eqs. \eqref{eq:scc-storage-set}-\eqref{eq:scc-storage-distance};
}
\Return $\tilde{\mathcal{D}}_m^k$;
\end{algorithm}

The detailed proof is given in Appendix \ref{proof:prop2}. Proposition \ref{prop:semantic-drift} reveals that two adjacent calibration rounds have a very similar semantic drift. Therefore, the semantic compensation step can be performed only once (i.e., on the first calibration round after each full synchronization) to save computational cost, and the backup parameters $\phi^{r_{\mathrm{bak}}}$ can be removed to release memory.

\vspace{1mm}
\textit{(b) Calibration.}
The compensated historical semantics ${\mathcal{Z}_m^k}^{*}$ will be concatenated with the latest semantics $\mathcal{S}_m^k(r)=h(\mathcal{D}_m^k(r),\phi_m^{k,r-1})$ to construct a calibration dataset $\tilde{\mathcal{D}}_m^k(r)$,
\begin{equation}\label{eq:scc-calibration-dataset}
\tilde{\mathcal{D}}_m^k(r)=\left[\mathcal{S}_m^k(r)\cup{\mathcal{Z}_m^k}^{*},\mathcal{Y}_m^k(r)\cup{\mathcal{Y}_m^k}^*\right],
\end{equation}
where $\mathcal{Y}_m^k(r)$ is the label space of $\mathcal{D}_m^k(r)$ and ${\mathcal{Y}_m^k}^*$ is the label set corresponding to ${\mathcal{Z}_m^k}^{*}$. 

When training the classifier, we freeze the parameters of feature extractor $\phi_m^{k,r-1}$, take $\tilde{\mathcal{D}}_m^k(r)$ as the training dataset (with labels), and use mini-batch SGD or Adam to optimize the classifier as follows,
\begin{equation}
\nonumber
\theta_m^{k,r}\gets\theta_m^{k,r-1}-\frac{\eta}{|\tilde{\mathcal{D}}_m^k(r)|}\nabla_{\theta}\mathcal{L}\left(g(\tilde{\mathcal{D}}_m^k(r), \theta_m^{k,r-1}), \mathcal{Y}_m^k\right).
\end{equation}

\vspace{1mm}
\textit{(c) Semantic Storage.}
The semantic database should be updated at every last round of calibration (i.e., the $(T-1)$-th round after each full synchronization). Looking back at the $T$ rounds just executed, we can construct a union set of candidate semantics $\mathcal{Q}$,
\begin{equation}\label{eq:scc-storage-set}
\mathcal{Q}={\mathcal{Z}_m^k}^{*}\cup\left(\bigcup\limits_{i=1}^{T}\mathcal{S}_m^k(r_i)\right),
\end{equation}
where $r_1$ is any full synchronization round and $r_2,\cdots,r_T$ are subsequent calibration rounds. We calculate the distance $\delta_i$ of each semantic feature $z_{i}\in\mathcal{Q}$ from its class mean $\mu_c^r$,
\begin{equation}\label{eq:scc-storage-distance}
\delta_{i}=|z_{i}-\mu_c^r|_{l_2}.
\end{equation}
The $Q$ semantics with the smallest value of $\delta_i$ are considered to be the most representative and are stored. Here, $Q$ is the maximum capacity of the semantic database.

Algorithm \ref{alg:lasp} gives the pseudo-code implementation of SCC that can be invoked by STP as a compatible module.

\subsection{LASP: Layer-wise Alternative Synchronization Protocol}
Generally, \NAME~consists of two types of synchronization modes: Full synchronization ($\texttt{FULL\_SYNC}$) and calibration ($\texttt{PART\_SYNC}$). Each full synchronization round is followed by $T-1$ calibration rounds. The main difference between them is that the full synchronization round trains the complete model parameters $\omega=\{\phi,\theta\}$, while the calibration round only trains the classifier $\theta$. Therefore, a concise and effective way to reduce communication costs is to synchronize only classifier parameters $\theta$ during calibration rounds, since the shallow layers $\phi$ are frozen and thus redundant.

\begin{figure}[H]
\centering
\includegraphics[width=.5\textwidth]{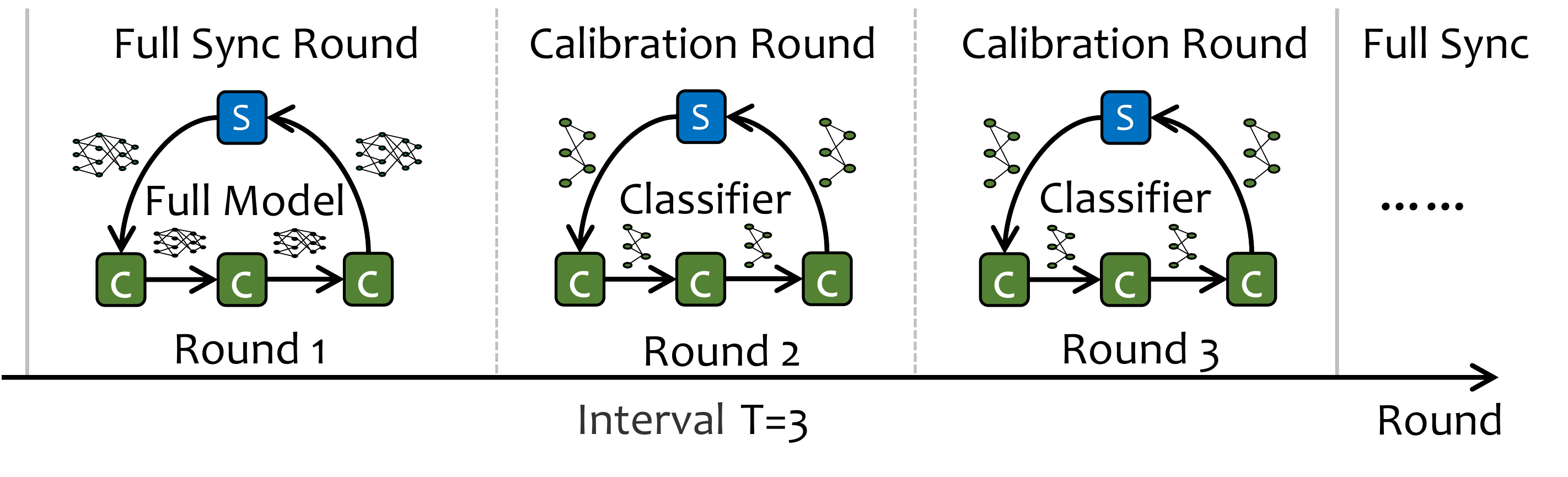}
\caption{An example of LASP workflow when the interval $T=3$.}
\label{fig:lasp}
\end{figure}

Based on this idea, we propose to synchronize the feature extractor $\phi$ and the classifier $\theta$ with different frequencies. An example is illustrated in Figure \ref{fig:lasp}. Assuming $r$ is a full synchronization round and the interval $T=3$, then $r+kT, \forall k\in\mathbb{Z}^*$ are also full synchronization rounds and $r+kT+1,r+kT+2, \forall k\in\mathbb{Z}^+$ are calibration rounds. In the full synchronization round, the complete model parameters $\omega=\{\phi,\theta\}$ are shared to enhance the semantic representation. Instead, in the calibration round, only classifier parameters $\theta$ are shared to reduce transfer bytes. We refer to this layer-wise and alternative synchronization protocol as LASP.

In summary, LASP has the following three advantages:
\begin{itemize}
\item LASP is easy to implement and well-compatible with existing synchronization algorithms.
\item LASP is a lossless approach that greatly reduces traffic by saving the transmission of redundant parameters. Table \ref{table:params-statistics} shows that 999\textperthousand~of traffic is saved on a typical CNN.
\item LASP helps to train a more generalized classifier since the classifier was found to be most sensitive to data heterogeneity\cite{luo2021no}.
\end{itemize}

\begin{table}[t]
  \caption{Statistics of parameters in a typical CNN model.}
  \small
  \label{table:params-statistics}
  \centering
  \renewcommand{\arraystretch}{1.2}
  \begin{tabular}{c|c|c|c|c|c}
    \hline
    & \textbf{Layer} & \textbf{Type} & \multicolumn{2}{c|}{\textbf{Num Params}} & \textbf{Ratio} \\
    \hline\hline
    \multirow{4}{*}{\makecell[c]{Feature\\Extractor}} & 1 & Conv & 0.8K & \multirow{4}{*}{\makecell[c]{Total:\\6.68M}} & \multirow{4}{*}{999\textperthousand} \\
    \cline{2-4} & 2 & Conv & 51K & & \\
    \cline{2-4} & 3 & FullyConnected & 6.4M & & \\
    \cline{2-4} & 4 & FullyConnected & 205K & & \\
    \hline
    \makecell[c]{Classifier} & 5 & FullyConnected & \multicolumn{2}{c|}{\textbf{6.3K}} & \textbf{1\textperthousand} \\
    \hline
  \end{tabular}
\end{table}
\section{Experiments and Results}\label{section:experiment}
\subsection{Experimental Setup}
We built the experimental platform on $K=368$ simulated devices and used the most common FEMNIST dataset\cite{caldas2018leaf} as the benchmark to evaluate the performance. FEMNIST is a non-i.i.d. dataset designed for FL, which contains 3,550 clients with 805,263 character samples divided according to a non-i.i.d. and non-uniform distribution. 

For resource-constrained industrial devices, we used a lightweight CNN network composed of two convolutional layers and two fully connected layers with a total of $\mathcal{M}=6.68\times 10^6$ parameters, where the classifier has about $\mathcal{M}_c=6.3\times 10^3$ parameters (see Table \ref{table:params-statistics}). 

The simulated devices use the standard mini-batch SGD to train their local models. Detailed hyperparameter settings are summarized in Table \ref{table:hyperparams}. The values of $T,f,\alpha,\beta$ will be further tuned in experiments to explore their impact on performance.

\begin{table}[t]
  \caption{Summary of main hyperparameter settings.}
  \small
  \label{table:hyperparams}
  \centering
  \renewcommand{\arraystretch}{1.2}
  \begin{tabular}{c|c|c}
    \hline
    \textbf{Definition} & \textbf{Symbol} & \textbf{Value} \\
    \hline\hline
    Number of devices & $K$ & 368 \\
    \hline
    Local learning rate & $\eta$ & 0.01 \\
    \hline
    Batch size & $b$ & 5 \\
    \hline
    Local epoch & $e$ & 1 \\
    \hline
    Maximum training rounds & $R$ & 500 \\
    \hline
    Interval of rounds & $T$ & 5 \\
    \hline
    Storage capacity of semantic database & $Q$ & 200 \\
    \hline
    Group sampling rate & $\kappa$ & 0.3 \\
    \hline
    Group growth function & $f(\cdot)$ & \texttt{LOG} \\
    \hline
    Coefficients for $f(\cdot)$ & $\alpha;\beta$ & 2; 10 \\
    \hline
    Size of streaming data batches & $n$ & 50 \\
    \hline
    Num of parameters for the full model & $\mathcal{M}$ & $6.68\times 10^6$ \\
    \hline
    Num of parameters for the classifier & $\mathcal{M}_c$ & $6.3\times 10^3$ \\
    \hline
  \end{tabular}
\end{table}

The experiments not only evaluate the performance of \NAME~on streaming non-i.i.d. data, but also the performance on traditional static non-i.i.d. data. 
Below we briefly introduce the difference between these two settings:
\begin{itemize}
\item \textbf{\textit{HFedMS-S}}: This setting enables only two features, that is, STP$+$ICG, and is suitable for processing static non-i.i.d. data. In this setup, the local data is fixed and the same dataset is used every round.
\item \textbf{\textit{HFedMS-D}}: This setting has all features enabled, that is, STP$+$ICG$+$SCC$+$LASP, and is suitable for processing dynamic streaming non-i.i.d. data. To generate streaming data, the client draws $n=50$ pieces of data from its local dataset each round and converts them into new samples using augmentation techniques.
\end{itemize}

Please note that the scheme in our previous work\cite{zeng2022heterogeneous} is the same as \NAME-S. The code implementation of this work is open available at \url{https://github.com/slz-ai/hfedms}.

\subsection{Results and Discussion}
\vspace{1mm}
\textbf{(a) Performance comparison with benchmarks.} First, we evaluate the overall performance (i.e., test accuracy and loss) and efficiency (i.e., training time and communication load) of our \NAME~under both streaming and static data settings, and compare it with several benchmarks to highlight its superiority. The benchmarks include FedAvg\cite{mcmahan2017communication}, FedProx\cite{li2020federated}, FedMMD\cite{yao2018two}, FedFusion\cite{yao2019towards}, IDA\cite{yeganeh2020inverse}, FedAdagrad, FedAdam, and FedYogi\cite{reddi2020adaptive}. The first is the vanilla algorithm for FL, and the rest are some state-of-the-art solutions proposed for non-i.i.d. data. All the benchmarks use the maximum training rounds $R=500$. For a fair comparison, our \NAME-D and \NAME-S also use $R=500$, that is, 100 full synchronization rounds and 400 calibration rounds in \NAME-D, and 500 full synchronization rounds in \NAME-S.

\begin{table}[t]
  \caption{Performance comparison with several benchmarks.}
  \small
  \label{table:hfedms-vs-others}
  \centering
  \renewcommand{\arraystretch}{1.2}
  \begin{tabular}{c|c|c|c|c} 
    \hline
    \multirow{2}{*}{\textbf{Algorithm}} & \multicolumn{2}{c|}{\textbf{Stream non-i.i.d. data}} & \multicolumn{2}{c}{\textbf{Static non-i.i.d. data}}  \\ 
    \cline{2-5}
    & \textbf{Acc} & \textbf{Loss} & \textbf{Acc} & \textbf{Loss} \\ 
    \hline\hline
    FedAvg & 70.1\% & 1.037 & 80.1\% & 0.602 \\
    FedProx & 70.3\% & 1.031 & 78.7\% & 0.633 \\
    FedMMD & 75.6\% & 0.821 & 81.7\% & 0.587 \\
    FedFusion & 72.6\% & 0.951 & 82.4\% & 0.554 \\
    IDA & 72.1\% & 1.003 & 82.0\% & 0.567 \\
    FedAdagrad & 77.4\% & 0.769 & 81.9\% & 0.582 \\
    FedAdam & 77.8\% & 0.730 & 82.1\% & 0.566 \\
    FedYogi & 75.8\% & 0.803 & 83.2\% & 0.543 \\ 
    \hline\hline
    \textbf{Algorithm} & \multicolumn{2}{c|}{\textbf{\NAME-D}} & \multicolumn{2}{c}{\textbf{\NAME-S}} \\
    \hline
    500 Rounds & \textbf{84.2\%} & \textbf{0.494} & \textbf{85.4\%} & \textbf{0.453} \\
    2141 Rounds & \textbf{85.9\%} & \textbf{0.401} & $\times$ & $\times$ \\
    \hline
  \end{tabular}
\end{table}

\begin{table}[t]
  \caption{Comparison of transfer bytes and training time.}
  \small
  \label{table:hfedms-efficiency}
  \centering
  \renewcommand{\arraystretch}{1.2}
  \begin{tabular}{c|c|c|c|c} 
    \hline
    \textbf{Algorithm} & \textbf{$\mathbf{T}$} & \textbf{$\mathbf{R}$} & \textbf{Transfer Bytes} & \textbf{Runtime} \\ 
    \hline\hline
    FedAvg & $\times$ & 490 & 2.629TB & 1262h \\ 
    \hline
    \NAME-S & 1 & 32 & 0.172TB & 82h \\
    \hline
    \multirow{4}{*}{\textbf{\NAME-D}} & 3 & 36 & 0.097TB & 42h \\
    & \textbf{5} & 34 & \textbf{0.056TB} & \textbf{25h} \\
    & 7 & 67 & 0.081TB & 35h \\
    & 9 & 81 & 0.073TB & 32h \\
    \hline
  \end{tabular}
\end{table}

Table \ref{table:hfedms-vs-others} gives the numerical results. It can be seen that \NAME~achieves significant performance improvements under both the stream and static non-i.i.d. data settings. For the static non-i.i.d. data, \NAME-S improves the benchmark accuracy by 2.2\%$\sim$6.7\%. Nevertheless, this is not all of \NAME, as \NAME-S does not have SCC and LASP enabled. The more advanced \NAME-D achieves a more significant breakthrough underlying streaming non-i.i.d. data. In this setting, \NAME-D outperforms the benchmarks by 6.4\%$\sim$14.1\% and maintains a satisfactory accuracy of 84.2\%, while the benchmark accuracy drops by 4.3\%$\sim$10\%. Nonetheless, \NAME-D has not converged to the optimum, and if we continue to train \NAME-D until convergence, we can obtain a better result of 85.9\%. These results demonstrate that \NAME-S is strongly robust to data heterogeneity, and its advanced version \NAME-D is also very suitable for processing dynamic streaming data.

Table \ref{table:hfedms-efficiency} compares the communication load and runtime efficiency when FedAvg and \NAME~reach the accuracy of 70\%. For \NAME, we take the interval $T=\{1,3,5,7,9\}$. It should be noted that \NAME-D with $T=1$ is actually \NAME-S. Eqs. \eqref{eq:hfedms-s-traffic}-\eqref{eq:hfedms-d-traffic} are used to count the total traffic,
\begin{align}
\text{\NAME-S:~}& 8\kappa K\mathcal{M}R, \label{eq:hfedms-s-traffic} \\
\text{\NAME-D:~}& 4\kappa K(3\mathcal{M}\lceil\frac{R}{T}\rceil+2(R-\lceil\frac{R}{T}\rceil)\mathcal{M}_c). \label{eq:hfedms-d-traffic}
\end{align}
Please note that Eqs. \eqref{eq:hfedms-s-traffic}-\eqref{eq:hfedms-d-traffic} do not take ICG into consideration since its traffic $4K\mathcal{F}\lceil R/T \rceil$ only accounts for a few hundred kilobytes and can be ignored. For the time cost, the runtime mainly consists of computational time and communication time, of which communication accounts for the majority due to the bandwidth bottleneck of LPWAN, while computation only takes a few minutes and can be ignored. The communication time was simulated using LTE Cat M2 with peak data rates of 4Mbps and 7Mbps for uplink and downlink, respectively. The results show that, regardless of the value of $T$, our \NAME~transfers far fewer bytes and runs in far less time than FedAvg, especially \NAME-D. The traffic was reduced by 90\%$\sim$98\% and the runtime was reduced by 93\%$\sim$98\%. This feature benefits from a cliff-like reduction in the training round $R$, about 83\%$\sim$93\% fewer.

To sum up, our \NAME~has four advantages: \textit{higher accuracy}, \textit{faster convergence}, \textit{less runtime}, and \textit{lower communication costs}, which makes it highly competitive in harsh industrial applications.
In follow-up experiments, we performed ablation experiments on \NAME~to help understand why it achieves such performance.

\vspace{1mm}
\textbf{(b) Effects of ICG and STP.}
We first define a benchmark algorithm for ablation experiments. This benchmark also divides clients into multiple groups and uses sequential training within each group and parallel training between groups. The difference is that the number and members of groups are fixed, like Astraea\cite{duan2019astraea}. Then, we apply ICG to this static benchmark for proper and fast grouping and refer it as ICG. Finally, STP is applied to make it dynamic, and it is now \NAME-S.

\begin{figure}[t]
\centering
\includegraphics[width=0.25\textwidth]{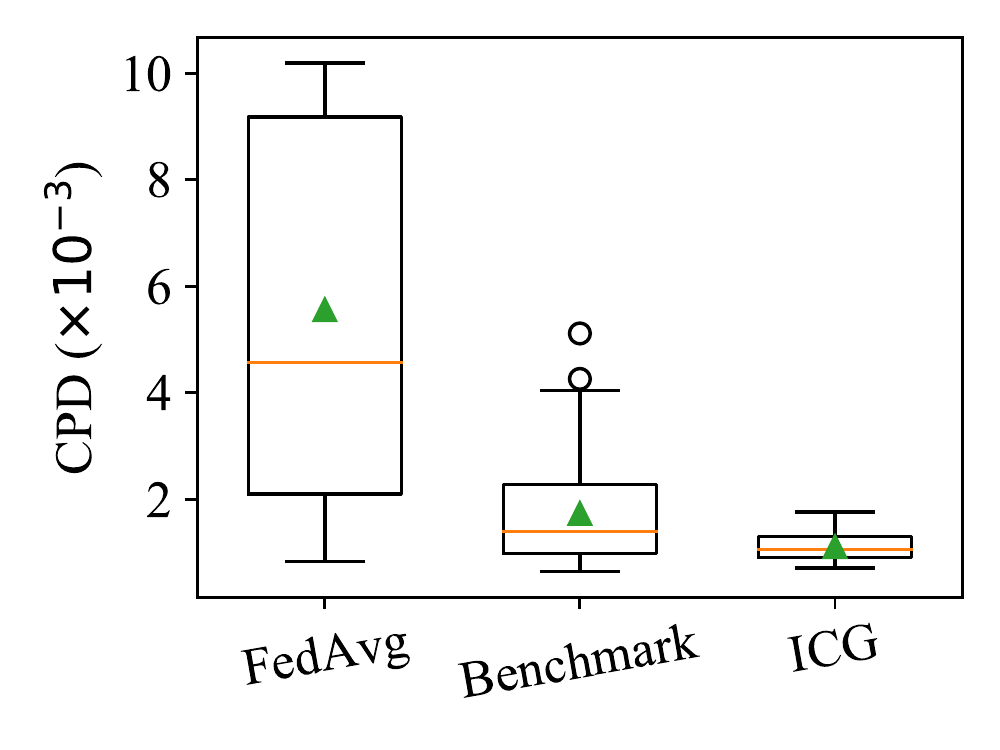}
\caption{Comparison of class probability distance.}
\label{fig:icg}
\end{figure}

Figure \ref{fig:icg} compares the class probability distances (CPDs) of FedAvg, Benchmark and ICG, where CPD is defined as the kernel two-sample estimation with Gaussian radial basis kernel $\mathcal{K}$,
\begin{align}
\nonumber
&\mathrm{CPD}(m_1,m_2)=\mathrm{MMD}^2(\mathcal{X}, \mathcal{Y}) \\
\nonumber
&= \mathbb{E}_{x,x'\sim\mathcal{X}}\left[\mathcal{K}(x,x')\right]-2\mathbb{E}_{x\sim\mathcal{X},y\sim\mathcal{Y}}\left[\mathcal{K}(x,y)\right]\\
\nonumber
&+\mathbb{E}_{y,y'\sim\mathcal{Y}}\left[\mathcal{K}(y,y')\right],
\end{align}
and $\mathcal{X}=\mathrm{norm}(\mathcal{V}_{m_1}),\mathcal{Y}=\mathrm{norm}(\mathcal{V}_{m_2})$ are normalized class probability distributions of two groups $\mathcal{G}_{m_1}, \mathcal{G}_{m_2}$. CPD quantifies the difference in the class probability distribution between two clients (or groups). In general, the smaller the CPD, the less heterogeneity between two clients (or groups), and thus the better the grouping strategy. In Figure \ref{fig:icg}, the CPD between every pair of clients (or groups) is counted, and the results show a remarkable reduction in CPD for our ICG, with the median value (with orange line) about 1/4 that of FedAvg and 2/5 that of Benchmark.

\begin{figure}[t]
\centering
\subfloat[]{\includegraphics[width=.245\textwidth]{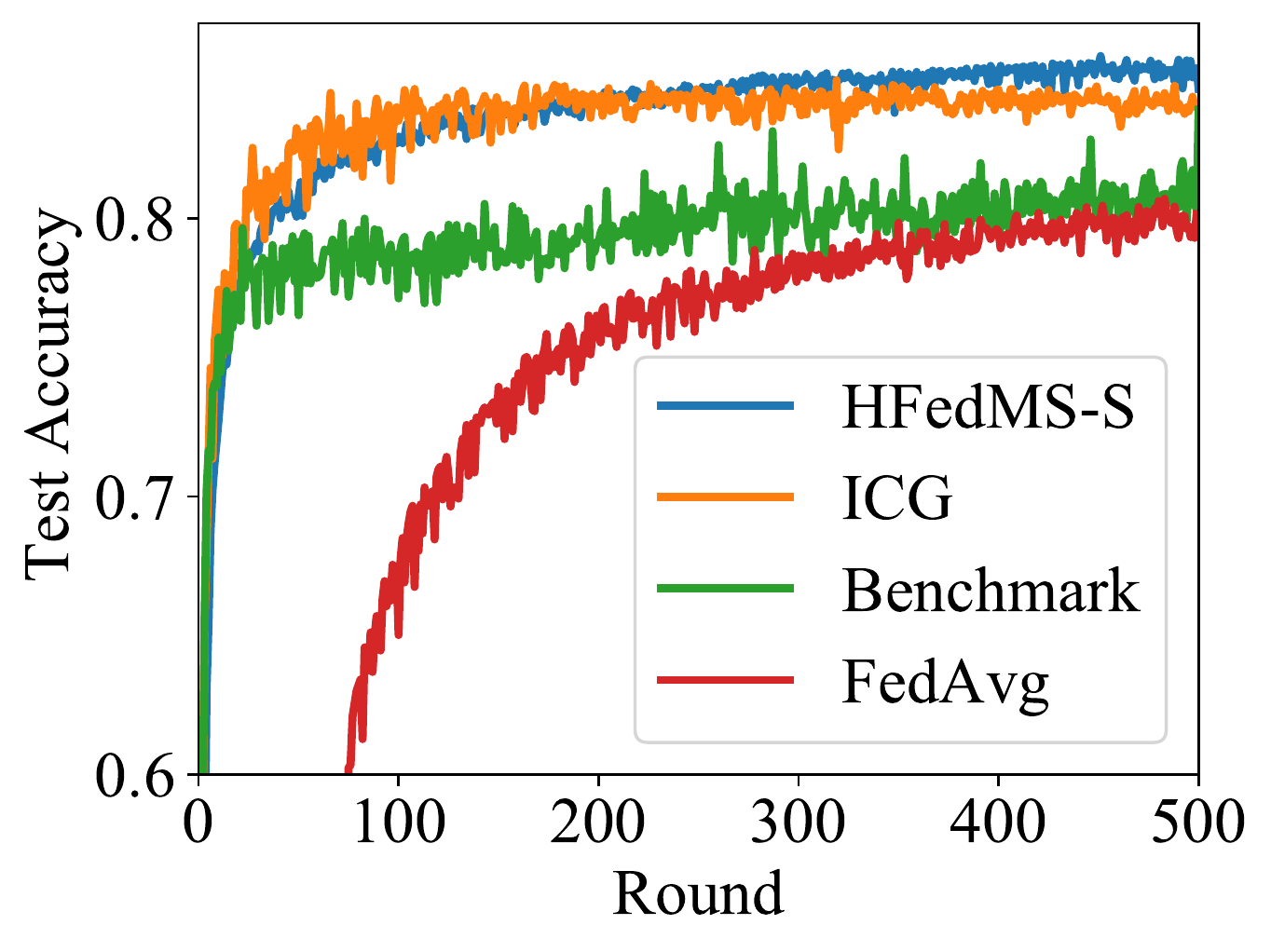}
\label{fig:acc-curve}}
\subfloat[]{\includegraphics[width=0.255\textwidth]{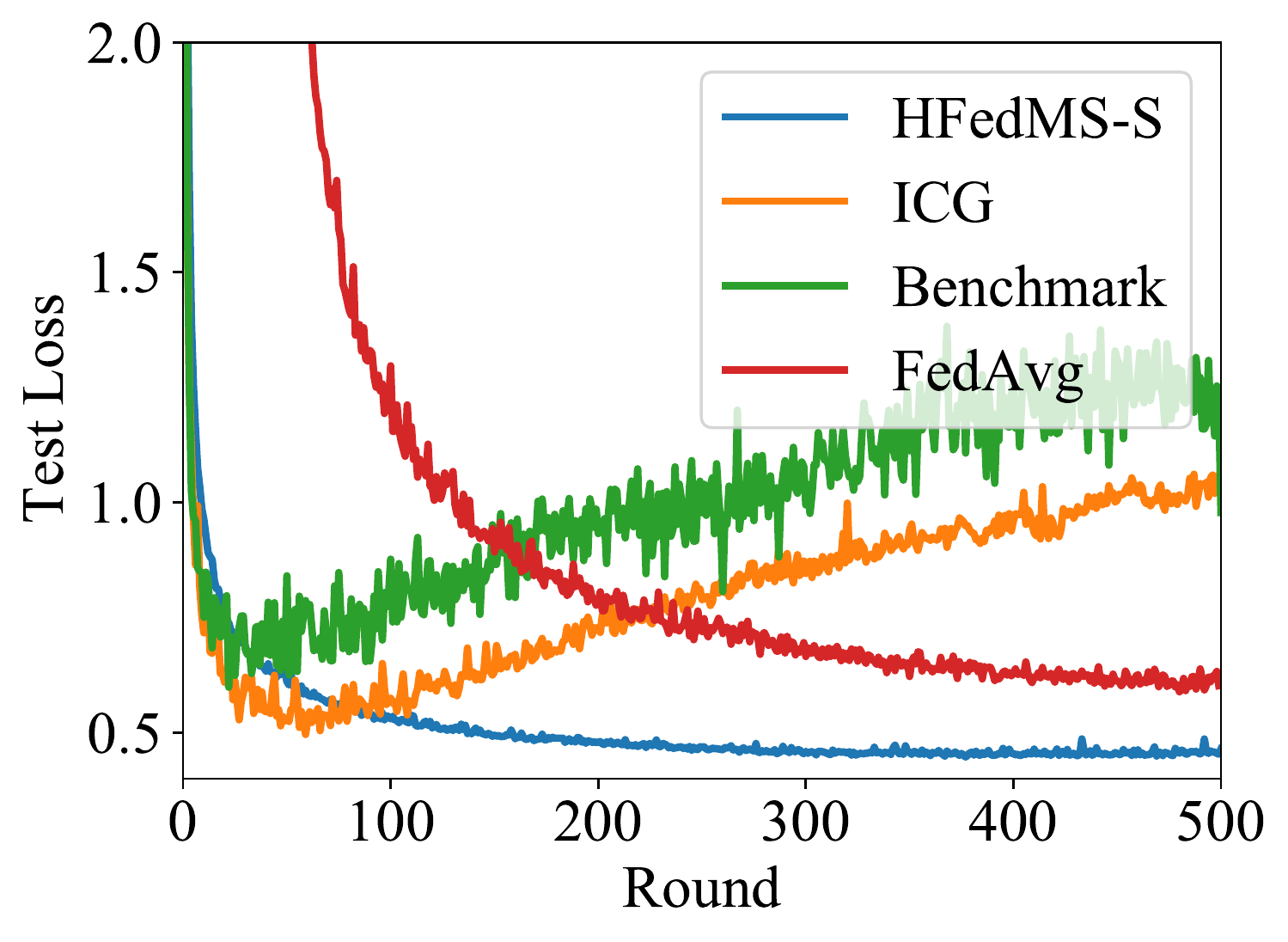}
\label{fig:loss-curve}}
\caption{Comparison of performance curves on static non-i.i.d. data.}
\label{fig:curve-on-static}
\end{figure}

\begin{figure*}
\centering
\subfloat[]{\includegraphics[width=.24\textwidth]{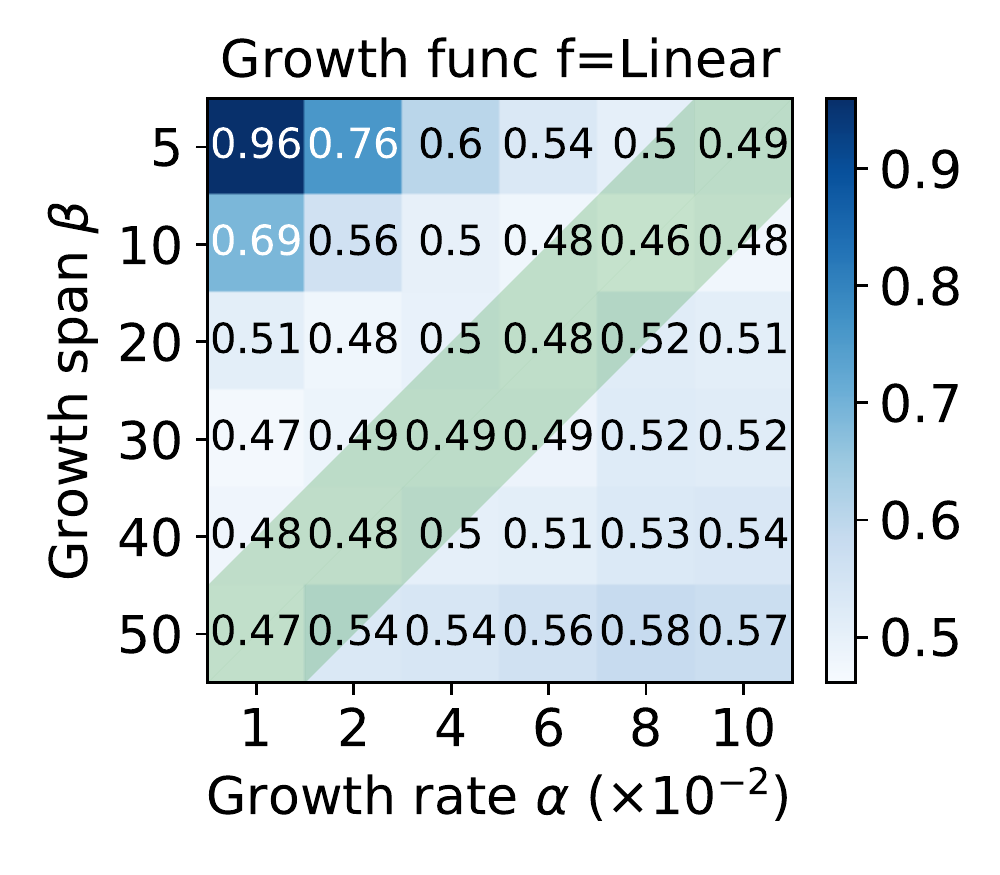}
\label{fig:hfedms-s-heatmap-linear}}
\subfloat[]{\includegraphics[width=.24\textwidth]{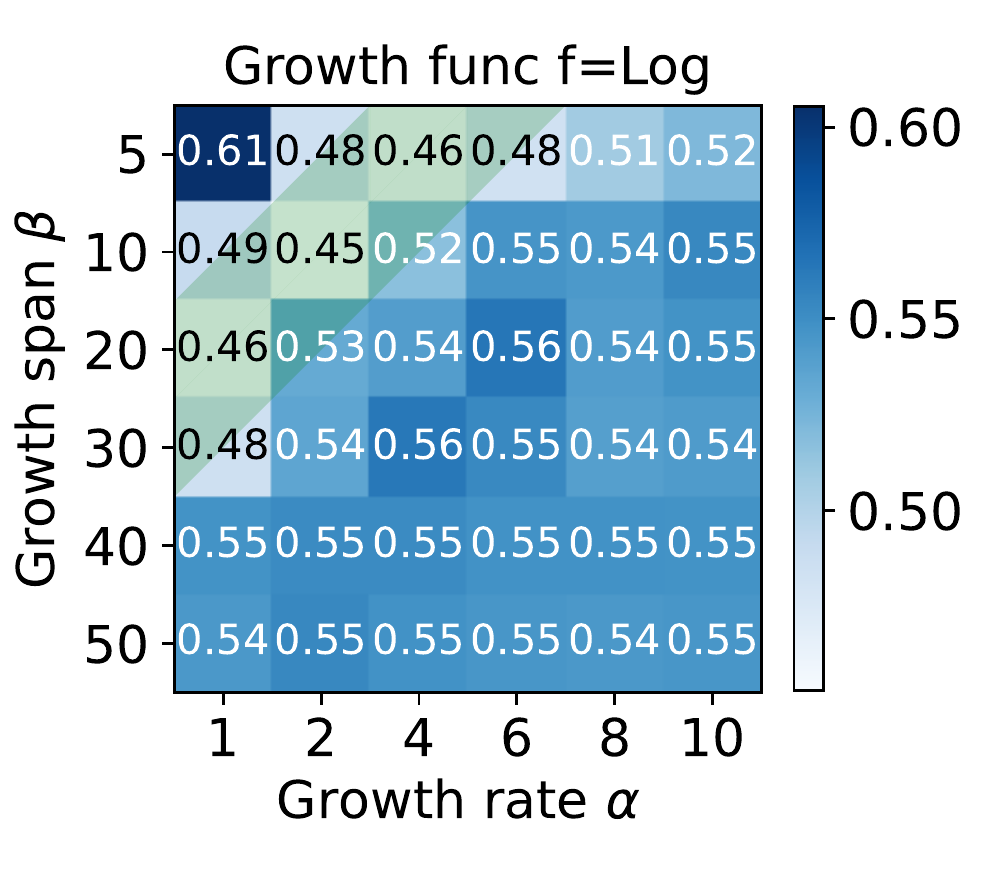}
\label{fig:hfedms-s-heatmap-log}}
\subfloat[]{\includegraphics[width=.24\textwidth]{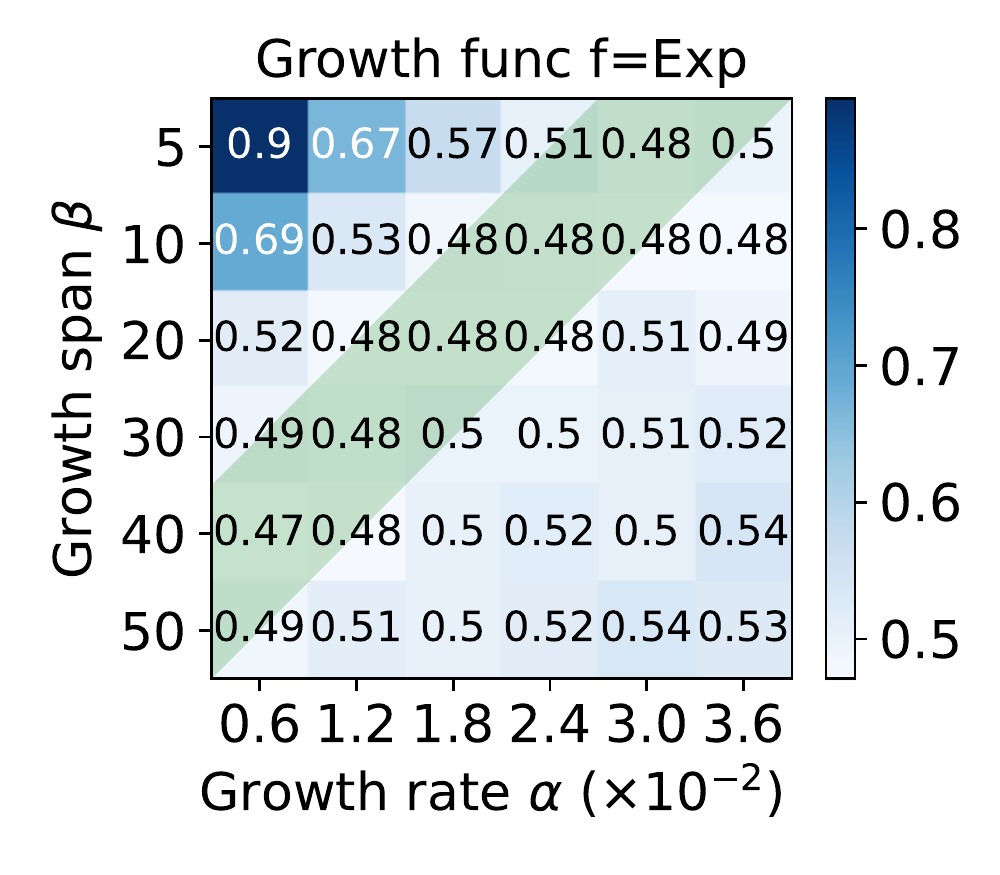}
\label{fig:hfedms-s-heatmap-exp}}
\subfloat[]{\includegraphics[width=.26\textwidth]{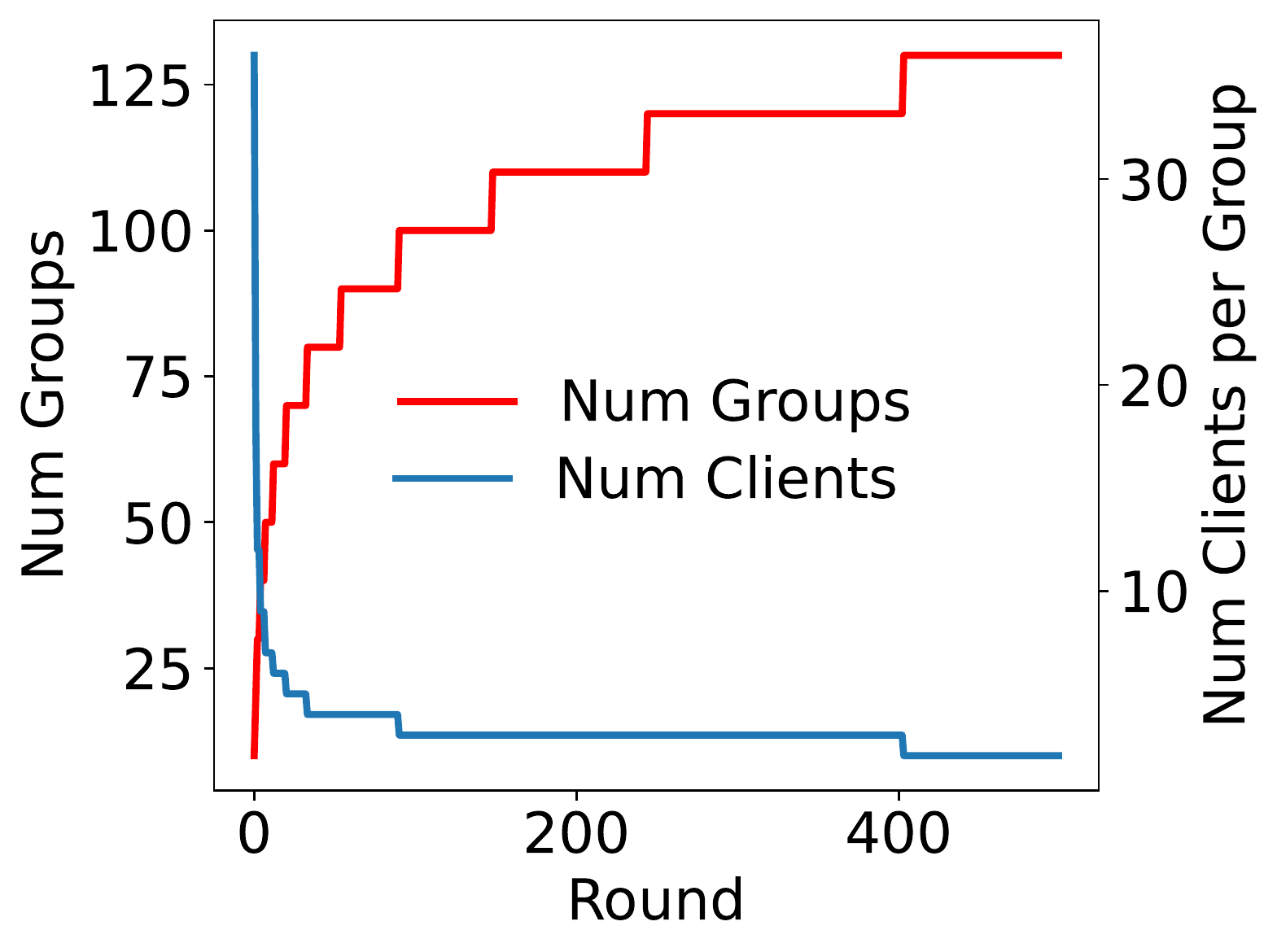}
\label{fig:hfedms-s-group-curve}}
\caption{Test loss heatmaps on (a) linear, (b) logarithmic, (c) exponential growth functions, and (d) curves of group status. The green area indicates the recommended range of $\alpha,\beta$ values.}
\label{fig:heatmap}
\end{figure*}

Figure \ref{fig:acc-curve} shows that native Benchmark converges significantly faster than vanilla FedAvg, while ICG further improves accuracy. This proves that constructing groups with homogeneous data distribution can indeed resist the negative effects of data heterogeneity. But surprisingly, Benchmark and ICG show diametrically opposite performances on the loss metric, that is, overfitting, as shown in Figure \ref{fig:loss-curve}. This is reasonable because of the static nature of Benchmark (i.e., the number and members of groups are fixed), which makes it possible for the FL model to learn some interfering information, such as the order in which clients perform sequential training. Therefore, \NAME-S was proposed to solve this problem, by introducing STP to make it dynamic. For example, regrouping clients, increasing the number of groups, and shuffling clients. The blue curves in Figures \ref{fig:acc-curve} and \ref{fig:loss-curve} perfectly overcome overfitting and converge to a higher accuracy of 85.4\%.

\vspace{1mm}
\textbf{(c) Effects of growth function and its coefficients.} 
The growth function $f(\cdot)$ and its coefficients $\alpha,\beta$ play an important role in STP. To investigate their effects, we conduct a grid search on $f=\{\texttt{LINEAR},\texttt{LOG},\texttt{EXP}\}$ and $\alpha,\beta$. Figures \ref{fig:hfedms-s-heatmap-linear}-\ref{fig:hfedms-s-heatmap-exp} gives the test loss heatmaps on \NAME-S and shows that the logarithmic growth function is most favored, where the loss value reaches the minimum 0.453 when $\alpha=2,\beta=10$. In fact, it is not good for the number of groups $M$ to grow too fast or too slowly. A slow increase in $M$ causes sequential training to dominate and leads to overfitting. Conversely, a rapid increase in $M$ causes \NAME-S to prematurely degenerate into FedAvg and suffer from data heterogeneity. Therefore, we recommend $\alpha\beta$ to be a moderate value, such as the green area in Figure \ref{fig:hfedms-s-heatmap-log}. In Figure \ref{fig:hfedms-s-group-curve}, we plot the actual curves of the number of groups and the number of clients within each group. An interesting finding is that, \NAME~prefers small groups with only 2$\sim$10 clients.

\begin{figure}[t]
\centering
\subfloat[]{\includegraphics[width=.255\textwidth]{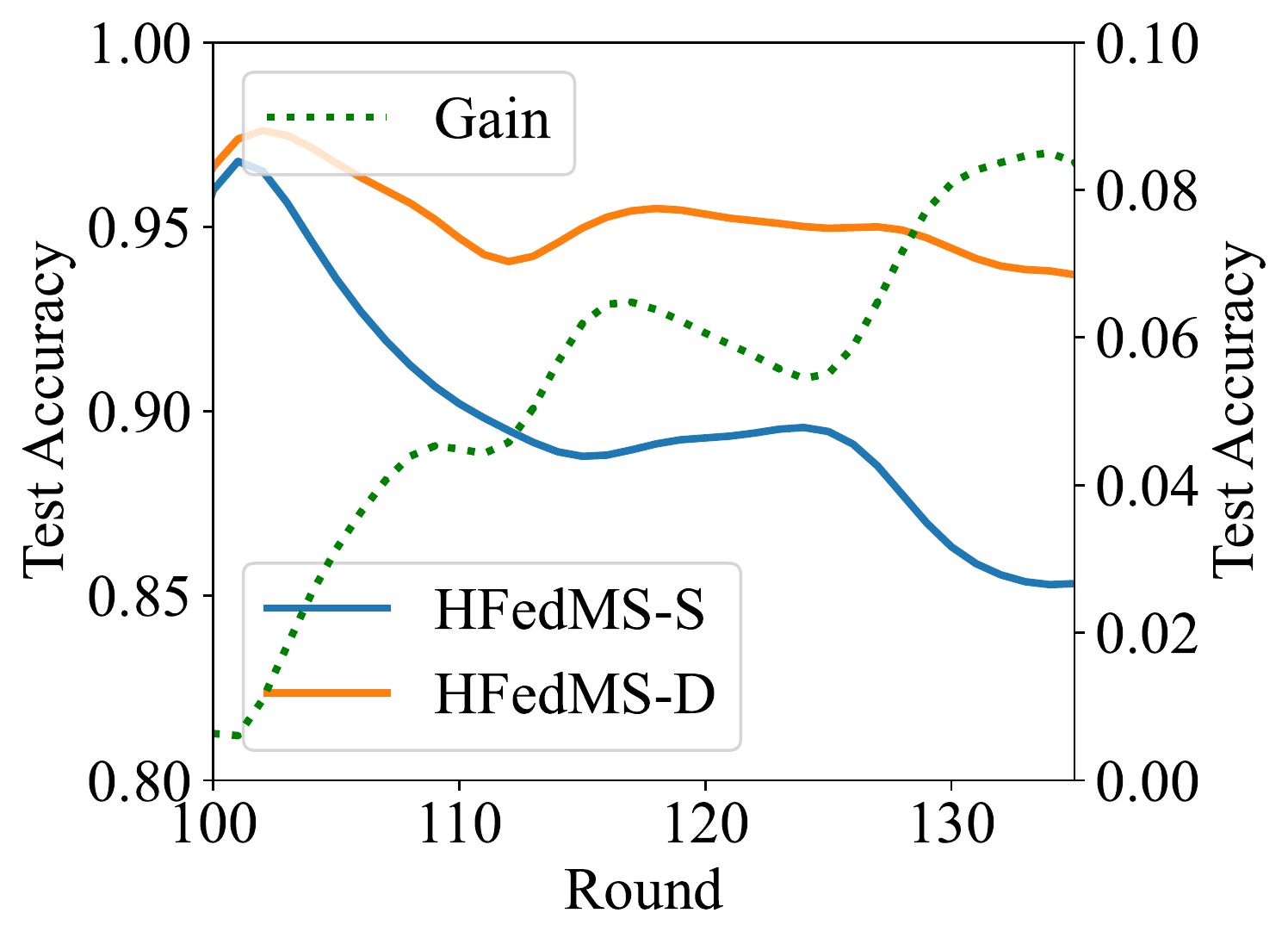}
\label{fig:scc-gain}}
\subfloat[]{\includegraphics[width=.245\textwidth]{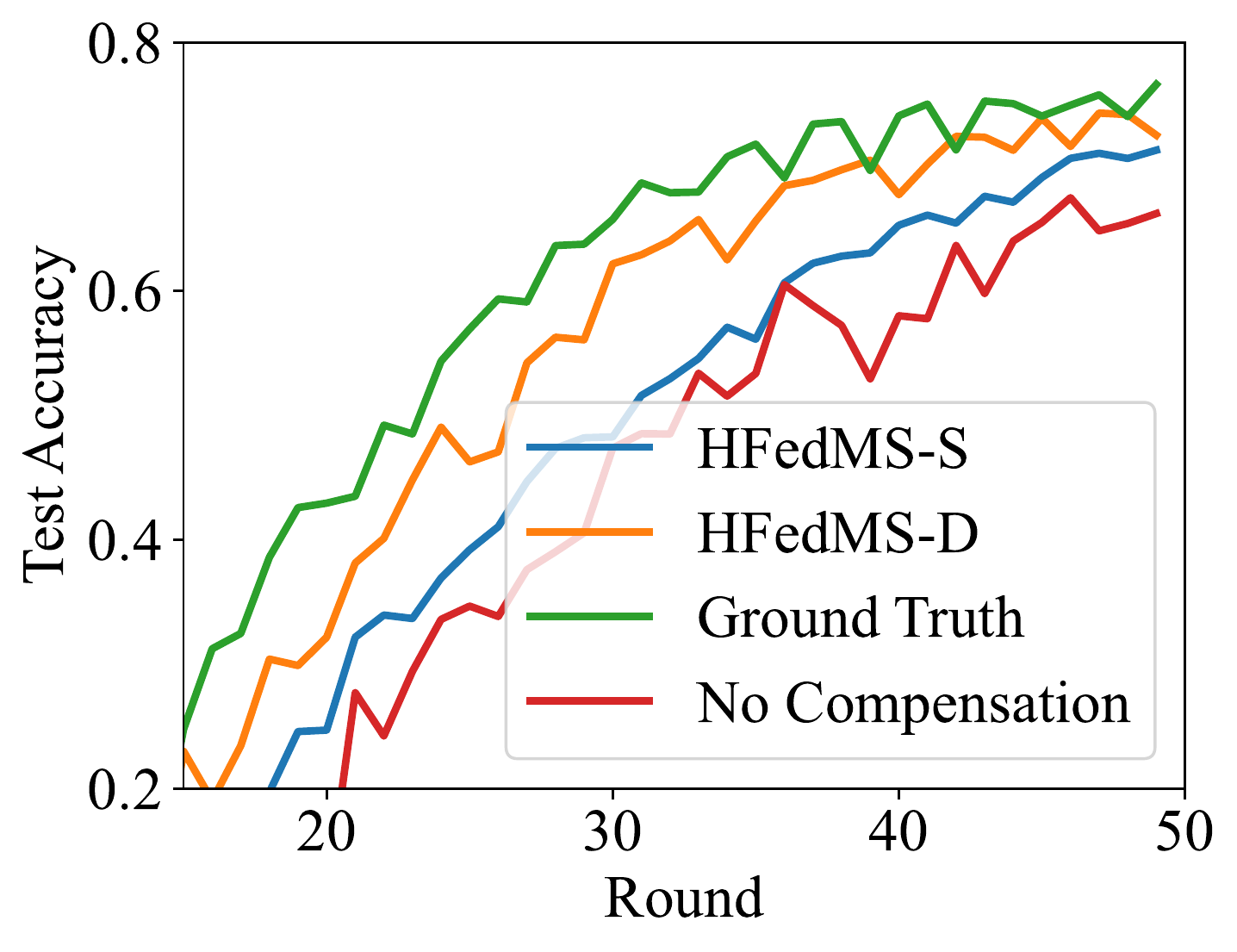}
\label{fig:scc-compensate}}
\caption{(a) The gain in accuracy and (b) accuracy curves before and after applying SCC.}
\label{fig:scc-curves}
\end{figure}

\begin{figure}[t]
\centering
\subfloat[]{\includegraphics[width=0.25\textwidth]{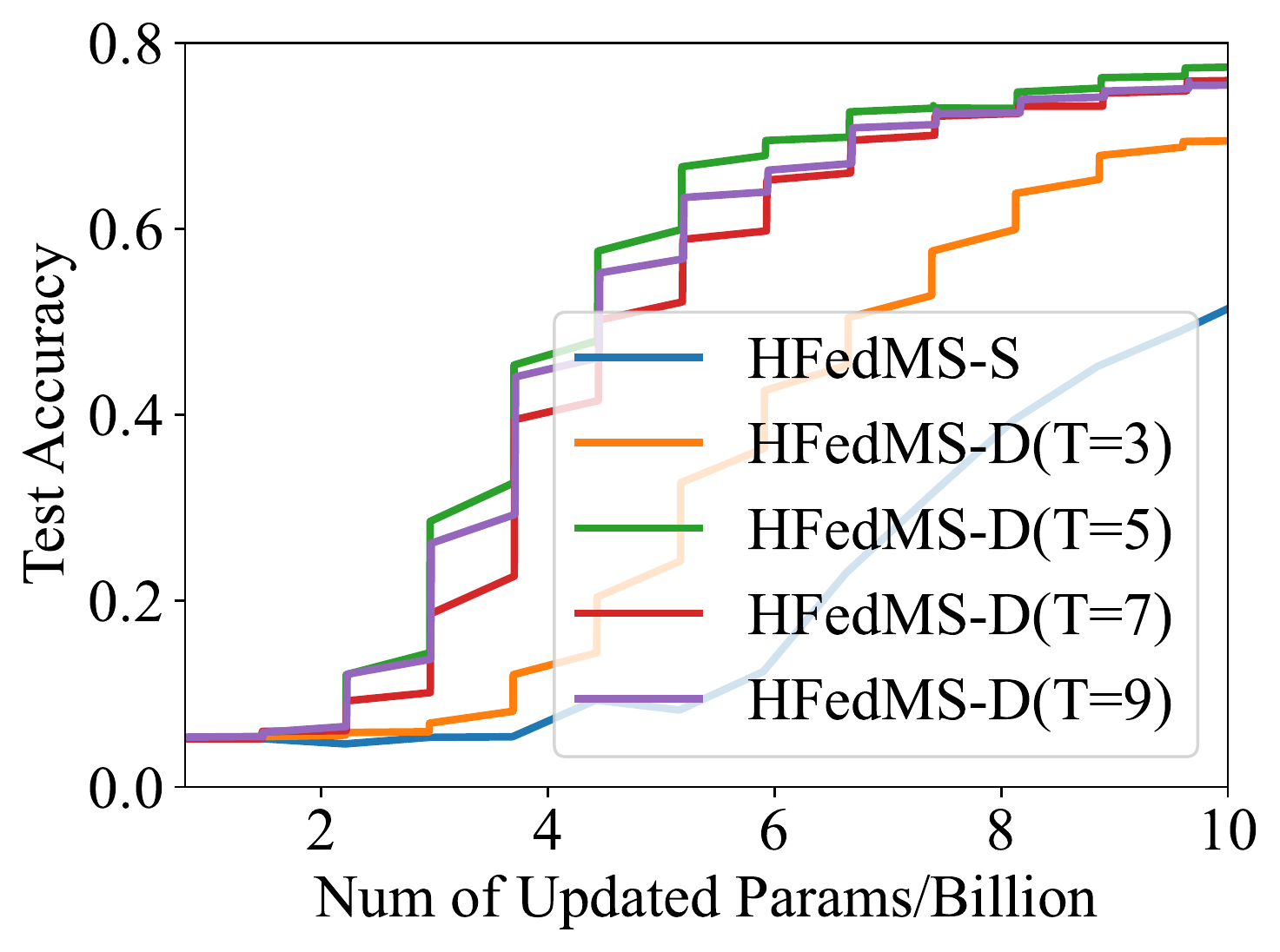}
\label{fig:lasp-acc}}
\subfloat[]{\includegraphics[width=0.25\textwidth]{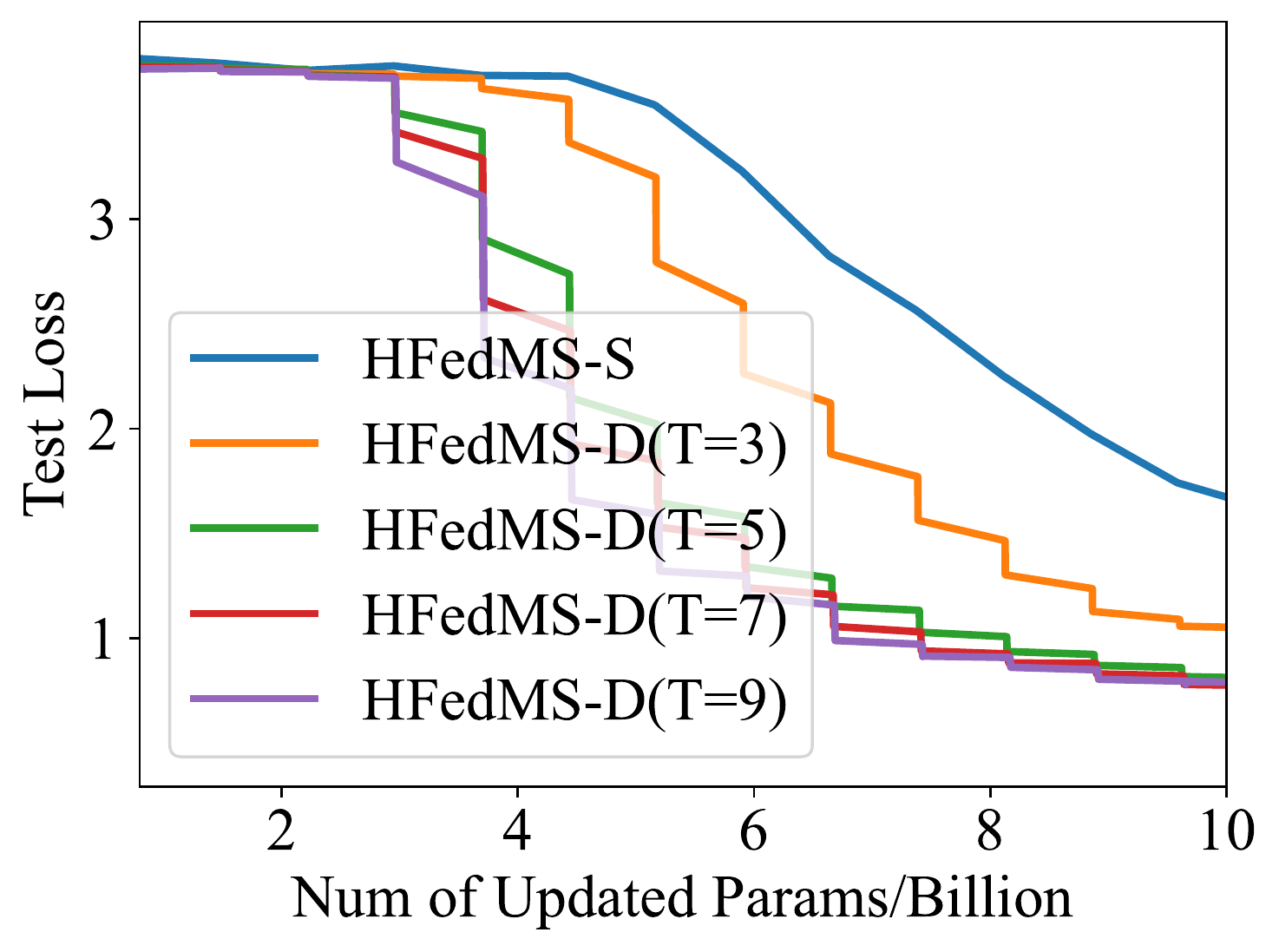}
\label{fig:lasp-loss}}
\caption{Comparison of (a) accuracy and (b) loss curves for different round intervals.}
\label{fig:lasp-curves}
\end{figure}

\vspace{1mm}
\textbf{(d) Effects of SCC.}
Although \NAME-S has shown exciting performance, it is still not enough to handle streaming non-i.i.d. data. In Table \ref{table:hfedms-efficiency}, \NAME-S and \NAME-D(T=5) achieve 70\% accuracy after similar rounds, however, this is an unfair comparison for \NAME-D as it runs only 7 full synchronization rounds, while the remaining 27 rounds of calibration only update the classifier. For a fair comparison, we use the total number of updated parameters as the x-axis to compare the convergence performance. From Figure \ref{fig:lasp-curves}, we can see that \NAME-D converges much faster on streaming non-i.i.d. data. This is expected as \NAME-S is not optimized for the learning forgetting problem, which is also our motivation to introduce SCC and propose a more advanced \NAME-D.

Figure \ref{fig:scc-gain} illustrates the phenomenon of learning forgetting. When we learn a batch of data in a certain round, in the subsequent rounds, the accuracy curve of \NAME-S on this data batch drops rapidly since these data no longer appear, this phenomenon is called learning forgetting. Instead, \NAME-D with SCC enabled maintains the accuracy well, and the accuracy gain achieved by resisting forgetting becomes more pronounced as the round progresses.

Figure \ref{fig:scc-compensate} also shows the effectiveness of SCC in improving overall performance. As expected, \NAME-D with SCC enabled converges faster and better than \NAME-S with SCC disabled, and it is also closest to the upper bound (in green line), which can store all the raw data and compensate the historical semantics accurately. Besides, we have an interesting finding that if we directly calibrate classifier with stored historical semantics without compensation (in red line), we would obtain worse result than that of \NAME-S with SCC disabled. This finding demonstrates the necessity and effectiveness of our compensation module.

\vspace{1mm}
\textbf{(e) Effects of LASP.} It is time-consuming to synchronize all parameters in all rounds in resource-constrained industrial networks. For example, in Table \ref{table:hfedms-efficiency}, FedAvg takes 1262 hours to achieve 70\% accuracy, and even \NAME-S takes 82 hours. To alleviate the communication bottleneck, LASP blocks the transmission of redundant parameters and saves 999\textperthousand~of transfer bytes in calibration rounds (see Table \ref{table:params-statistics}). For overall performance, as shown in Table \ref{table:hfedms-efficiency}, LASP-enabled \NAME-D achieves up to 67\% savings in transfer bytes compared to \NAME-S, and thus reduces runtime by nearly 70\%. For the setting of the round interval $T$, the overall performance reaches the best when $T=5$, at this time, the FL system has the highest accuracy, the shortest runtime, the fewest rounds, and the lowest communication cost. 

\vspace{1mm}
\textbf{(f) \NAME-S vs \NAME-D.} Figure \ref{fig:lasp-curves} shows the state-of-the-art of \NAME-D in terms of convergence performance when dealing with streaming non-i.i.d. data. For \NAME-D, the curve contains and alternates sharp- and slow-rising curves. The slow-rising curves indicate that the algorithm runs in full synchronization rounds, and the sharp-rising curves run in calibration rounds. In general, \NAME-D works much better than \NAME-S, with significantly faster convergence and higher accuracy. This is intuitively due to the jump in accuracy in calibration rounds (the sharp-rising curves), and it also proves that SCC plays a pivotal role in improving performance.
\section{Conclusions}
Federated Learning, as one of the enabling technologies laying the foundation for Metaverse, provides powerful collaborative data analysis tools with privacy protection guarantees. Facing severe constraints and key challenges of incorporating FL into Industrial Metaverse, that is, data heterogeneity, learning forgetting, and limited bandwidth, this paper proposed a high-performance and efficient FL system, named \NAME, to simultaneously address these three challenges. This system uses a new concept of STP training mode to achieve dynamic collaboration between devices, and a fast ICG algorithm to group devices and minimize heterogeneity. Then, SCC is proposed to address learning forgetting induced by streaming data, which compensates for the forgotten knowledge by fusing compressed historical data semantics and calibrates only classifier parameters. Finally, the system runs in LASP protocol to block the transmission of redundant parameters and reduce communication costs. Extensive experiments were conducted on the static and streamed non-i.i.d. FEMNIST dataset. Numerical results clearly show that \NAME~achieves state-of-the-art performance compared to eight benchmarks in both settings, with significantly a higher accuracy, faster convergence, less runtime, and lower communication costs. These advantages demonstrate the advanced performance of our \NAME~system on streaming non-i.i.d. data.

% Appendix
% \begin{appendices}
% \input{appendix.tex}
% \end{appendices}

% References
\bibliographystyle{IEEEtran}
\bibliography{ref}


% \end{IEEEbiography}


%\newpage



%\vfill


\end{document}

% Appendix
\begin{appendices}

\section{Proof of Proposition 1}\label{proof:prop1}
\begin{proof}
The expectation of cluster centroid $C^m$ can be deduced by
\begin{align}
\nonumber
\mathbb{E}[C^m]&=\mathbb{E}[\frac{1}{L}\sum_{l=1}^{L}\mathcal{V}^m_l] \\
\nonumber
&=\mathbb{E}[\frac{1}{L}\sum_{l=1}^{L}(\mathcal{V}^m_l-C_l+C_l)] \\
\nonumber
&=\mathbb{E}[\frac{1}{L}\sum_{l=1}^{L}(\mathcal{V}^m_l-C_l)+\frac{1}{L}\sum_{l=1}^{L}{C_l}] \\
\nonumber
&=\frac{1}{L}\mathbb{E}[\epsilon^m]+C_{\mathrm{global}} =C_{\mathrm{global}},
\end{align}
This implies that the group centroid and the global centroid are coincident in expectation, and the error is bounded by
\begin{align}
\nonumber
\|C^m-C_\mathrm{global}\|_2^2&=\|\frac{1}{L}\sum_{l=1}^{L}{\mathcal{V}_l^m}-\frac{1}{L}\sum_{l=1}^{L}{C_l}\|_2^2 \\
\nonumber
&=\frac{1}{L^2}\|\sum_{l=1}^{L}{(\mathcal{V}_l^m-C_l)}\|_2^2 \\
\nonumber
&\le\frac{1}{L^2}\sum_{l=1}^{L}\|\mathcal{V}_l^m-C_l\|_2^2 =\frac{1}{L^2}\sum_{l=1}^{L}{\sigma_l^2}.
\end{align}
\end{proof}

\vspace{5mm}
\section{Proof of Proposition 2}\label{proof:prop2}
\begin{proof}
First, we have
\begin{align}
\nonumber
\mu^{r_1}&=\mathbb{E}_{x_i}[h(\mathcal{D}_m^k({r_1}),\phi^{{r_1}-1})]=\mathbb{E}_{x_i}[\mathcal{D}_m^k({r_1})]\cdot\Phi^{{r_1}-1}, \\
\nonumber
\mu^{r_2}&=\mathbb{E}_{x_i}[h(\mathcal{D}_m^k({r_2}),\phi^{{r_2}-1})]=\mathbb{E}_{x_i}[\mathcal{D}_m^k({r_2})]\cdot\Phi^{{r_2}-1}, \\
\nonumber
\mu^{\tilde{r}}_1&=\mathbb{E}_{x_i}[h(\mathcal{D}_m^k(r_1),\phi^{\tilde{r}})]=\mathbb{E}_{x_i}[\mathcal{D}_m^k(r_1)]\cdot\Phi^{\tilde{r}}, \\
\nonumber
\mu^{\tilde{r}}_2&=\mathbb{E}_{x_i}[h(\mathcal{D}_m^k(r_2),\phi^{\tilde{r}})]=\mathbb{E}_{x_i}[\mathcal{D}_m^k(r_2)]\cdot\Phi^{\tilde{r}}.
\end{align}
The semantic drift can be defined as
\begin{align}
\nonumber
\Delta^{r_1}&=\mu^{r_1}-\mu_1^{\tilde{r}}=\mathbb{E}_{x_i}[\mathcal{D}_m^k(r_1)]\cdot(\Phi^{{r_1}-1}-\Phi^{\tilde{r}}), \\
\nonumber
\Delta^{r_2}&=\mu^{r_2}-\mu_2^{\tilde{r}}=\mathbb{E}_{x_i}[\mathcal{D}_m^k(r_2)]\cdot(\Phi^{{r_2}-1}-\Phi^{\tilde{r}}).
\end{align}
To simplify the representation, we make $\mathbb{E}_{x_i}[\mathcal{D}_m^k(r_2)]$ aliased as $\Gamma$ and have $\mathbb{E}_{x_i}[\mathcal{D}_m^k(r_1)]=\Gamma+\Xi$ according to Assumption 2. Then we have $\Delta^{r_1}-\Delta^{r_2}$ as 
\begin{align}
\nonumber
\Delta^{r_1}-\Delta^{r_2}&=(\Gamma+\Xi)\cdot(\Phi^{r_1-1}-\Phi^{\tilde{r}})-\Gamma\cdot(\Phi^{r_2-1}-\Phi^{\tilde{r}}) \\
\nonumber
&=\Gamma\cdot(\Phi^{r_1-1}-\Phi^{r_2-1})+\Xi\cdot(\Phi^{r_1-1}-\Phi^{\tilde{r}}) \\
\nonumber
&\overset{1}{=}\Xi\cdot(\Phi^r-\Phi^{\tilde{r}}),
\end{align}
``$\overset{1}{=}$" holds because $\Phi^r=\Phi^{r_1-1}=\Phi^{r_2-1}$. Therefore, the expectation and covariance of $\Delta^{r_1}-\Delta^{r_2}$ over any round $r_1$ and $r_2$ are
\begin{align}
\nonumber
\mathbb{E}_r[\Delta^{r_1}-\Delta^{r_2}]&=\mathbb{E}_r[\Xi]\cdot(\Phi^r-\Phi^{\tilde{r}})\overset{2}{=}\boldsymbol{0}, \\
\nonumber
\mathrm{Cov}(\Delta^{r_1}-\Delta^{r_2})&=\mathbb{E}_r[(\Delta^{r_1}-\Delta^{r_2}-\mathbb{E}_r[\Delta^{r_1}-\Delta^{r_2}])^2] \\
\nonumber
&=\mathbb{E}_r[(\Xi\cdot(\Phi^r-\Phi^{\tilde{r}}))^2] \\
\nonumber
&=(\Phi^r-\Phi^{\tilde{r}})^T\cdot\mathbb{E}_r[\Xi^T\cdot\Xi]\cdot(\Phi^r-\Phi^{\tilde{r}}) \\
\nonumber
&\overset{3}{\rightarrow}\boldsymbol{0},
\end{align}
where ``$\overset{2}{=}$" holds because $\mathbb{E}_r[\Xi]=\boldsymbol{0}$, and ``$\overset{3}{\rightarrow}$" holds because $|\Xi|\rightarrow 0$ makes $\Xi^T\cdot\Xi\rightarrow\boldsymbol{0}$.
\end{proof}

\end{appendices}